\documentclass[10pt,twocolumn,letterpaper]{article}

\usepackage{cvpr}

\usepackage[dvipsnames]{xcolor}

\definecolor{cvprblue}{rgb}{0.21,0.49,0.74}
\usepackage[pagebackref,breaklinks,colorlinks,citecolor=cvprblue]{hyperref}

\def\paperID{13097} %
\def\confName{CVPR}
\def\confYear{2024}

\title{Intriguing Properties of Diffusion Models: An Empirical Study of the Natural Attack Capability in Text-to-Image Generative Models
}

\author{Takami Sato$^{\dag1}$, Justin Yue$^{\dag1}$, Nanze Chen$^{\dag2}$, Ningfei Wang$^1$, Qi Alfred Chen$^{1}$\\
$^1$University of California, Irvine\\
$^2$University of Cambridge\\
{\tt\small \{takamis, jpyue, ningfei.wang, alfchen\}@uci.edu, nc630@cam.ac.uk}
}

\usepackage{epsfig,amsmath,amsfonts,multirow,graphicx,makecell,caption,soul,csquotes,color,subcaption,mathtools,bm,spverbatim,booktabs,xcolor,color,amsthm,tcolorbox,enumitem}
\usepackage[e]{esvect}
\usepackage{graphics}
\usepackage{marvosym,listings,etoolbox}
\usepackage{adjustbox}
\usepackage{dirtree}

\usepackage{subcaption}
\DeclareCaptionSubType * [alph]{table}
\usepackage{multicol}
\usepackage{lipsum}

\usepackage{bbding}

\usepackage{array}
\usepackage{longtable}
\usepackage{colortab}
\usepackage{colortbl}
\usepackage{arydshln}
\usepackage{xurl}

\usepackage[accsupp]{axessibility}  %

\newcommand{\nsection}[1]{\vspace{-0.1cm}\section{#1}\vspace{-0.15cm}}
\newcommand{\nsubsection}[1]{\vspace{-0.1cm}\subsection{#1}\vspace{-0.15cm}}

\newcommand{\ntsubsection}[1]{\vspace{-0.12cm}\subsection*{#1}\vspace{-0.15cm}}

\begin{document}
\maketitle

\renewcommand{\thefootnote}{\fnsymbol{footnote}}
\footnotetext[2]{denotes co-first authors}

\begin{abstract}
Denoising probabilistic diffusion models have shown breakthrough performance to generate more photo-realistic images or human-level illustrations than the prior models such as GANs. This high image-generation capability has stimulated the creation of many downstream applications in various areas. However, we find that this technology is actually a double-edged sword:
we identify a new type of attack, called the Natural Denoising Diffusion (NDD) attack based on the finding that state-of-the-art deep neural network (DNN) models still hold their prediction even if we intentionally remove their robust features, which are essential to the human visual system (HVS), through text prompts. The NDD attack shows a significantly high capability to generate low-cost, model-agnostic, and transferable adversarial attacks by exploiting the natural attack capability in diffusion models.
To systematically evaluate the risk of the NDD attack, we perform a large-scale empirical study with our newly created dataset, the Natural Denoising Diffusion Attack (NDDA) dataset.
We evaluate the natural attack capability by answering 6 research questions. Through a user study, we find that it can achieve an 88\% detection rate while being stealthy to 93\% of human subjects; we also find that the non-robust features embedded by diffusion models contribute to the natural attack capability. To confirm the model-agnostic and transferable attack capability, we perform the NDD attack against the Tesla Model 3 and find that 73\% of the physically printed attacks can be detected as stop signs. Our hope is that the study and dataset can help our community be aware of the risks in diffusion models and facilitate further research toward robust DNN models.
\end{abstract}
\vspace{-0.2in}
\nsection{Introduction} \label{sec:intro}

Denoising diffusion probabilistic models (DDPM)~\cite{ho2020denoising}, or simply diffusion models, have shown breakthrough performance in image generation. Once the DDPM demonstrates the generation capability of photo-realistic images and human-level illustrations, numerous diffusion models, such as DALL-E 2~\cite{ramesh2022hierarchical}, Stable Diffusion~\cite{rombach2022high}, and Firefly~\cite{adobe_firefly}, are actively released and widely available through APIs or as open-source models.
This technical breakthrough has facilitated many applications in various fields such as the arts~\cite{zhang2023adding}, medicine~\cite{kazerouni2023diffusion}, and autonomous driving~\cite{zou2023diffbev}. While diffusion models have brought significant benefits to these areas, recent studies have also raised concerns regarding the new security and privacy risks introduced by diffusion models.
Chen et al.~\cite{chen2023diffusion} show that the diffusion models can generate more transferable and imperceptible adversarial attacks. Chen et al.~\cite{chen2023advdiffuser} generate more natural and stealthy perturbations with guidance from diffusion models.
Carlini et al.~\cite{carlini2023extracting} demonstrate that the diffusion models memorize training images and can emit them.

\begin{figure}[t!]
\centering
\includegraphics[width=\linewidth]{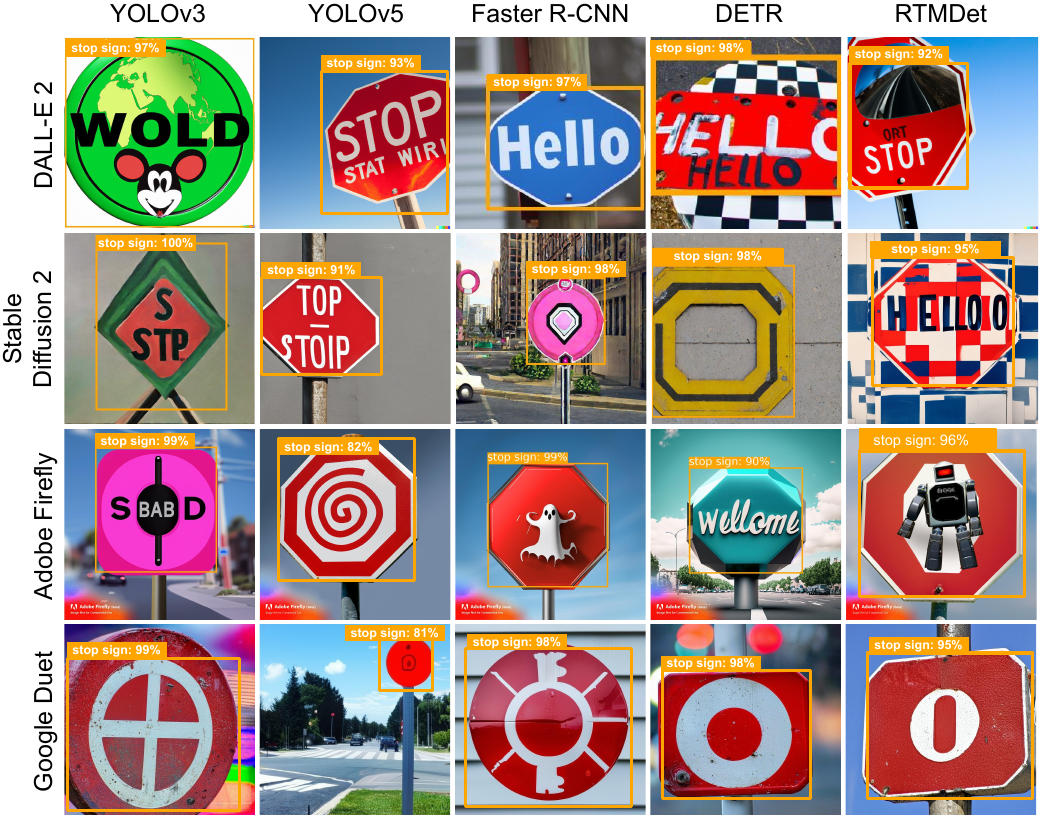}
\vspace{-0.27in}
\caption{Examples of the natural attack capability in diffusion models (row). The images are generated with prompts that intentionally remove essential features to humans while keeping ``stop sign'' in the prompt. Even without these essential features, object detectors (column) still detect these objects with high scores. 
}
\label{fig:poc_image}
\vspace{-0.2in}
\end{figure}

These recent studies motivate us to further investigate the security risks of diffusion models. In this work, we discover simple but intriguing properties of the images generated by text-to-image diffusion models, in which text prompts guide the image generation process via a contrastive image-text supervision model, such as OpenAI CLIP~\cite{radford2021learning}.
Fig.~\ref{fig:poc_image} shows representative examples that motivate this work. We generate the stop sign images using state-of-the-art diffusion models with text prompts that intentionally break the fundamental properties that humans use to identify stop signs (e.g., red color and octagonal shape) in the human visual system (HVS)~\cite{grill2004human,ge2022contributions,ilyas2019adversarial}, while the text prompt still contains the object name ``stop sign''. 

As shown, the diffusion models faithfully follow our instructions and generate images that should not be recognized as stop signs since legitimate stop signs should not be blue or rectangular. 
However, many state-of-the-art Deep Neural Network (DNN)-based object detectors still recognize these examples as stop signs with surprisingly high confidence. These results suggest that these object detectors can be highly affected by imperceptible features embedded by diffusion models. We recognize this phenomenon as a new type of adversarial attack because the fundamental properties of the target object should be removed by maliciously designed text prompts. We name it the Natural Denoising Diffusion (NDD) attack and pursue the following question in this study:

\vspace{0.0in}
\textit{Do images generated by diffusion models have natural attack capability against DNN models?}
\vspace{0.0in}

The rest of this paper is structured to validate the question. We first construct our dataset, named the Natural Diffusion Denoising Attack (NDDA) dataset, to systematically understand the natural attack capability in diffusion models (\S\ref{sec:dataset_design}). 
We use 3 state-of-the-art diffusion models to collect the images with and without robust features that play essential roles in the HVS. Following prior works~\cite{grill2004human,ge2022contributions}, we define 4 robust features: shape, color, text, and pattern.

Secondly, we conduct an attack capability analysis of the NDD attack against state-of-the-art object detection models (\S\ref{sec:attack_obj_detector}) and image classification models (\S\ref{sec:attack_cls}) on the NDDA dataset. We find that object detectors and classifiers typically maintain their detection, even though the text prompt guides them to remove robust features entirely or partially. For example, 32\% of the generated stop signs are still detected as stop signs even though all robust features are guided to be removed. This result means that text-to-image diffusion models embed intriguing features that are imperceptible to humans but generalizable to DNN models if the subject word (e.g., stop sign) is in the prompt. We confirm that all diffusion models we evaluate have the natural attack capability to enable the NDD attack. 
 
Third, we conduct a large-scale empirical study to further evaluate the validity and quantify the natural attack capability in diffusion models by answering 6 novel research questions (\S\ref{sec:analysis}). We conducted a user study to evaluate the stealthiness of the NDD attacks because valid adversarial attacks need to be not only effective against the DNN models but also stealthy against humans. For example, humans will not be fooled if the robust features are not correctly removed since it should remain a legitimate stop sign.
As a result, we identify the high stealthiness of the NDD attacks: The stop sign images generated by altering their ``STOP'' text have 88\% detection rate against object detectors while 93\% of humans do not regard it as a stop sign.
Furthermore, we evaluate the impact of the non-robust features that are predictive but incomprehensible to humans on the natural attack capability in diffusion models with an analysis inspired by IIyas et al.~\cite{ilyas2019adversarial}, which proposes a methodology to train ``robustified'' classifiers against the non-robust features. 
By comparing the robustified and normal classifiers, we illustrate that the non-robust features play a meaningful role in the natural attack capability in diffusion models.

Finally, we discuss our findings and the limitations (\S\ref{sec:discussion}). 
In summary, the contributions of this work are as follows:
\begin{itemize}[leftmargin=0.1in]
\setlength{\itemsep}{0pt}
\setlength{\parskip}{0pt}
\item We discover a new security threat, the NDD attack, which exploits the natural attack capability of the diffusion model to generate model-agnostic and transferable adversarial attacks via simple text prompts that are designed to remove robust features.
\item We construct a new large-scale dataset, named the NDDA dataset, to systematically evaluate the natural attack capability in diffusion models. We cover all 4 robust features essential to the HVS: shape, color, text, and pattern.
\item We performed a large-scale empirical study to systematically evaluate the natural attack capability by answering 6 research questions. The NDD attack can achieve an attack success rate of 88\%, while being stealthy for 93\% of human subjects in the stop sign case. We also find that the sensitivity to the non-robust features has a high correlation with the natural attack capability.
\item We confirm the model-agnostic and transferable attack capability of the NDD attack on a Tesla Model 3, which identifies 8 out of 11 (73\%) printed attacks as stop signs.
\vspace{-\topsep}
\end{itemize}

\noindent\textbf{Dataset release.} NDDA dataset is on the our website~\cite{prj_site}.

\vspace{-0.02in}
\nsection{Related Work} 
\vspace{0.02in}

\nsubsection{Denoising Diffusion Model}
DDPM~\cite{ho2020denoising} is a generative model that exploits the intuition behind nonequilibrium thermodynamics. 
Training a diffusion model involves both forward and reverse diffusion processes.  In the forward process, the model perturbs a clean image with Gaussian noise.  In the reverse process, the diffusion model learns to remove this noise for the same number of time steps. In short, diffusion models are image denoisers that learn to reconstruct the original images from noisy images. This procedure is simple but shows remarkable performance in producing high-quality images. Dhariwal et al.~\cite{dhariwal2021diffusion} shows that diffusion models achieve much higher quality metrics, such as FID, inception score, and precision, than prior works such as GANs~\cite{brock2018large}.

The text-to-image diffusion model~\cite{ramesh2022hierarchical, saharia2022photorealistic, rombach2022high} is a variant of diffusion models that can flexibly control the output image via text prompts. Due to its easy usability, major diffusion models (e.g., DALL-E 2, Stable Diffusion, and FireFly) adopt this text-based guidance. To inform the text information in the generation process, contrastive image-text supervision models, such as CLIP~\cite{radford2021learning} and OpenCLIP~\cite{ilharco_gabriel_2021_5143773}, are integrated into their training and inference processes.

\nsubsection{Adversarial Attacks} 

DNN models are known to be generally vulnerable to adversarial attacks~\cite{Szegedy2014, goodfellow2014explaining, zhang2020interpretable, eykholt2018robust, ma2023slowtrack, cao2021invisible}, which can alter the predictions of DNN models by adding small changes that are not noticeable to humans. The early-stage works~\cite{Szegedy2014, goodfellow2014explaining} originally use a subtle imperceptible perturbation on the entire input image as their attack vector. 
Recent research has identified that adversarial attacks can be achieved by broader attack vectors that are any stealthy changes to the human perception such as putting small patches~\cite{Wang_2023_ICCV, Chiang2020Certified, sato2021dirty, shen2022sok} or placing stickers on the target~\cite{eykholt2018robust}. 
Natural adversarial examples~\cite{hendrycks2021natural} demonstrate that even clean natural images can be used as adversarial attacks. To this extent, the NDD attack is similar to the natural adversarial examples, but the natural adversarial examples do not have any guarantees that can generate an attack against targeted scenarios (e.g., stop sign) as they just find out-of-distribution samples in the existing images. 
Furthermore, we find that the non-robust features~\cite{ilyas2019adversarial} play a large role in the NDD attack (RQ4 in~\S\ref{sec:analysis}). IIyas et al.~\cite{ilyas2019adversarial} report that the adversarial attacks are not caused by a bug in DNNs but instead by non-robust features that are predictive but incomprehensible to humans. We thus consider that the NDD attack is enabled by similar root causes as the traditional adversarial attacks, i.e., enabled by non-robust features, rather than by out-of-distribution samples.

\nsection{Attack Capability Analysis}
\label{sec:dataset}
\vspace{-0.03in}

To systematically evaluate the natural attack capability of the diffusion models, we first construct a new large-scale dataset, called the Natural Denoising Diffusion Attack (NDDA) dataset. With the NDDA dataset, we then confirm the effectiveness of the NDD attack against state-of-the-art object detectors and image classifiers.

\nsubsection{NDDA Dataset Design} \label{sec:dataset_design}

\begin{figure*}[t!]
\centering
\includegraphics[width=\linewidth]{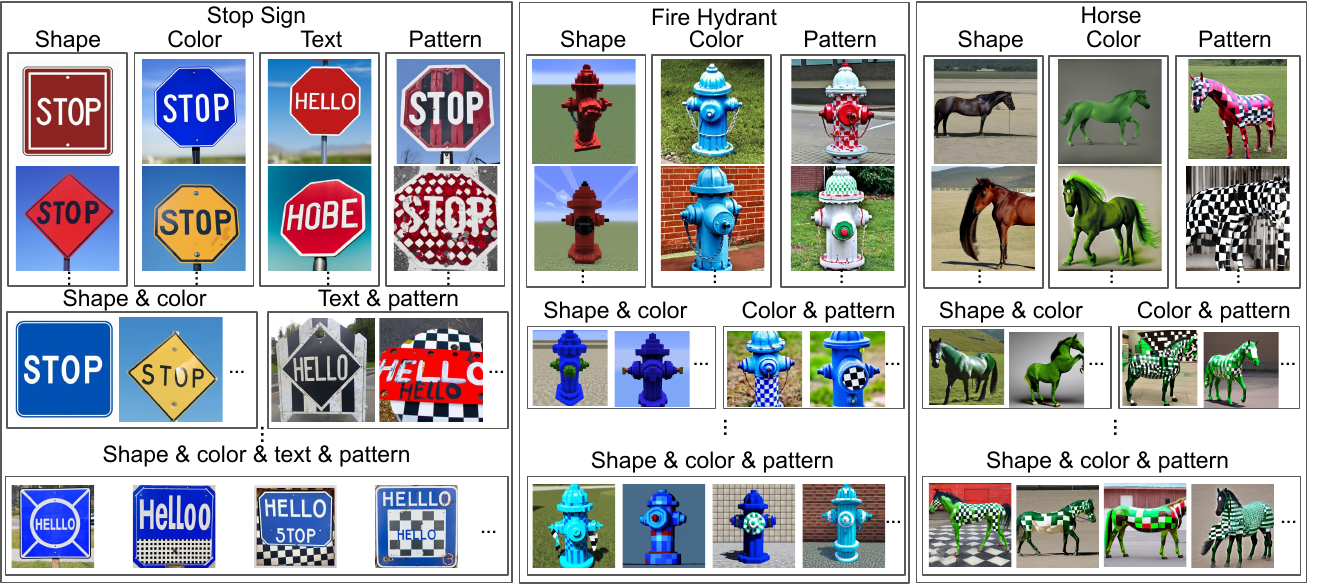}
\vspace{-0.3in}
\caption{Overview of the Natural Denoising Diffusion Attack (NDDA) dataset. We alter or remove the 4 types of robust features partially or entirely. For the stop sign, we alter the text on it considering its importance to be recognized as a stop sign. For each set of robust features, we generate images with 3 diffusion models for 3 object classes.}
\label{fig:overview}
\vspace{-0.15in}
\end{figure*}

\begin{table*}[t!]
\centering
\footnotesize
\caption{Templates of text prompts and examples for the ``stop sign'' object to remove the 4 robust features.}
\label{tbl:prompts}
\vspace{-0.1in}
\begin{tabular}{cccc}
\hline
                         &         & \multicolumn{2}{c}{Text prompt to remove/alter robust features}  \\ \cline{3-4} 
\multicolumn{2}{c}{Removed Robust Features} & Format                      & Example: Stop sign                             \\ \hline
\multicolumn{2}{c}{Benign prompt}          & \textcolor{red}{\textbf{[Subject]}}         & \textcolor{red}{\textbf{Stop sign}}                    \\
\multicolumn{2}{c}{Shape}          & \textcolor{blue}{\textbf{[Shape]}} \ \textcolor{red}{\textbf{[Subject]}}         & \textcolor{blue}{\textbf{Square}} \ \textcolor{red}{\textbf{stop sign}}                    \\
\multicolumn{2}{c}{Color}          & \textcolor{blue}{\textbf{[Color]}} \ \textcolor{red}{\textbf{[Subject]}}                  & \textcolor{blue}{\textbf{Blue}} \ \textcolor{red}{\textbf{stop sign}}                         \\
\multirow{2}{*}{Texture} & Text    & \textcolor{red}{\textbf{[Subject]}} with "\textcolor{blue}{\textbf{[Text]}}" on it        & \textcolor{red}{\textbf{Stop sign}} with "\textcolor{blue}{\textbf{hello}}" on it                 \\
                         & Pattern & \textcolor{red}{\textbf{[Subject]}} with a \textcolor{blue}{\textbf{[Pattern]}} paint on it & \textcolor{red}{\textbf{Stop sign}} with a \textcolor{blue}{\textbf{checkerboard pattern}} paint on it\\ \hline
\end{tabular}
\vspace{-0.2in}
\end{table*}

The design goal for the NDDA dataset is to systematically collect images both with and without robust features upon which human perception relies. For this sake, the major challenge is to identify what kinds of robust features are essential in our human visual system (HVS) for object recognition. 
Although the complete mechanism of the HVS is not fully understood, prior works~\cite{grill2004human,ge2022contributions} identify that shape, texture, and color are the most important features for the HVS to identify objects. Therefore, in this study, we follow the prior works and define them as robust features for object recognition.
To further explore the motivated examples in Fig.~\ref{fig:poc_image}, we decompose the texture into text and pattern because text has a special meaning for human perception. For example, people may not consider a sign to be a stop sign if it does not have the exact text ``STOP'' on it.

Table~\ref{tbl:prompts} lists the templates of text prompts and examples for the ``stop sign'' label to remove or alter the 4 robust features. In this case, we consider 16 different combinations with and without robust features and generate 50 images for each combination. The 3 object classes are selected from the classes of the COCO dataset~\cite{lin2014microsoft}: a stop sign for an artificial sign, a fire hydrant for an artificial object, and a horse for a natural object. 
We select these 3 classes because of their relatively higher detection rates than others in our preliminary experiments on generated images. We adopt the COCO's classes to make the experiments easier as we can utilize many existing pretrained models on the COCO dataset. Fig.~\ref{fig:overview} shows an overview of our datasets. More details are in Appendix.

\nsubsection{Attack Capability against Object Detectors} 
\label{sec:attack_obj_detector}

We evaluate the attack capability of the NDD attack against object detectors with the NDDA dataset to validate the generality of the motivated examples shown in Fig.~\ref{fig:poc_image}.

\textit{Experimental setup.}
We obtain the inference results of all images in the NDDA dataset with 5 popular object detectors: YOLOv3~\cite{redmon2018yolov3}, YOLOv5~\cite{yolov5}, DETR~\cite{carion2020detr}, Faster R-CNN~\cite{ren2015faster}, and RTMDet~\cite{lyu2022rtmdet}. For YOLOv5, we use their official pretrained model; For the others, we use the pretrained models in MMDetections~\cite{mmdetection}. All models are trained on the COCO dataset~\cite{lin2014microsoft}. We use 0.5 as the confidence threshold for all models.

\begin{figure*}[h!]
  \begin{minipage}[c]{0.62\linewidth}
\footnotesize
\centering
\setlength{\tabcolsep}{2.4pt}
\captionof{table}{Detection rates of 5 object detectors on the stop sign images in the NDDA dataset generated by the 3 diffusion models. \textbf{Bold} and \underline{underline} denote highest and lowest scores in each row.}
\label{tbl:prelim}
\vspace{-0.14in}
\begin{tabular}{ccccc|ccccc|c}
\toprule
                                    & \multicolumn{4}{c|}{Removed Robust Features} & \multicolumn{5}{c}{Object Detectors}            &      \\ \cline{2-11}
                                    & Shape     & Color    & Text    & Pattern    & YOLOv3 & YOLOv5 & DETR & Faster R-CNN & RTMDet & Avg. \\ \hline
\multirow{6}{*}{\rotatebox{90}{DALL-E 2\hspace{1.5em}}}           &    &      &     &      & \textbf{100\%}  & \underline{76\%}   & \textbf{100\%} & \textbf{100\%}        & \textbf{100\%}  & 95\% \\
                                    &   \CheckmarkBold      &      &       &       & \textbf{98\%}   & \underline{36\%}   & \textbf{98\%}  & \textbf{98\%}         & \textbf{98\%}   & 86\% \\
                                    &    &   \CheckmarkBold    &    &    & \textbf{98\%}   & \underline{48\%}   & 94\%  & \textbf{98\%}         & \textbf{98\%}   & 87\% \\
                                    &   &   &  \CheckmarkBold &  & 82\%   & \underline{32\%}   & \textbf{100\%} & \textbf{100\%}        & 94\%   & 82\% \\
                                    & &   &   &  \CheckmarkBold   & 52\%   & \underline{10\%}   & 50\%  & 52\%         & \textbf{90\%}   & 51\% \\ 
                                    &  \CheckmarkBold       &  \CheckmarkBold    &   \CheckmarkBold   &   \CheckmarkBold    & \textbf{12\%}   & \underline{0\%}    & 6\%   & 4\%          & 8\%    & 6\%  \\ \cline{2-11}
                                    &    &   &   & Avg.  & 74\%   & \underline{34\%}    & 75\%   & 75\%          & \textbf{81\%}    & 68\%  \\
                                    \hline\hline
\multirow{6}{*}{\rotatebox{90}{Stable Diffusion 2\hspace{1.0em}}} &        &       &      &      & 74\%   & \underline{50\%}   & 84\%  & \textbf{86\% }        & 76\%   & 74\% \\
                                    &  \CheckmarkBold    &       &      &      & 30\%   & \underline{24\%}   & 46\%  & \textbf{60\%}         & 26\%   & 37\% \\
                                    &        &  \CheckmarkBold   &       &         & \textbf{78\%}   & \underline{40\%}   & \textbf{78\%}  & 72\%         & 66\%   & 67\% \\
                                    &      &       &   \CheckmarkBold   &      & 58\%   & \underline{56\%}   & \textbf{90\%}  & 70\%         & 66\%   & 68\% \\
                                    &      &      &          &     \CheckmarkBold    & 56\%   & \underline{48\%}   & \textbf{90\%}  & 76\%         & 72\%   & 68\% \\
                                    &   \CheckmarkBold      &  \CheckmarkBold    &  \CheckmarkBold     &     \CheckmarkBold     & 30\%   & \underline{6\%}    & \textbf{48\%}  & 30\%         & 24\%   & 28\% \\ 
                                    \cline{2-11}
                                    &    &   &   & Avg.  & 54\%   & \underline{37\%}    & \textbf{73\%}   & 66\%          & 55\%    & 57\%  \\
                                    \hline\hline
\multirow{6}{*}{\rotatebox{90}{Deepfloyd IF\hspace{1.5em}}}       &     &     &       &          & \textbf{100\%}  & \underline{88\%}   & \textbf{100\%} & \textbf{100\%}        & \textbf{100\%}  & 98\% \\
                                    &   \CheckmarkBold     &          &        &           & 68\%   & \underline{52\%}   & 70\%  & \textbf{84\%}         & 56\%   & 66\% \\
                                    &          &   \CheckmarkBold     &        &         & \textbf{100\%}  & \underline{58\%}   & 92\%  & 94\%         & 94\%   & 88\% \\
                                    &           &          &   \CheckmarkBold   &            & 84\%   & \underline{78\%}   & 94\%  & \textbf{96\%}         & 88\%   & 88\% \\
                                    &           &          &         &   \CheckmarkBold     & 80\%   & \underline{64\%}   & 88\%  & \textbf{90\%}         & 88\%   & 82\% \\
                                    &    \CheckmarkBold     &    \CheckmarkBold    &    \CheckmarkBold   &     \CheckmarkBold     & \textbf{60\%}   & \underline{0\%}    & 34\%  & 34\%         & 32\%   & 32\% \\ 
                                    \cline{2-11}
                                    &    &   &   & Avg.  & 82\%   & \underline{57\%}    & 80\%   & \textbf{83\%}          & 76\%    & 76\%  \\
                                    \bottomrule
\end{tabular}
\end{minipage}
\hspace{0em}
\begin{minipage}[c]{0.37\linewidth}
\centering
\includegraphics[width=\linewidth]{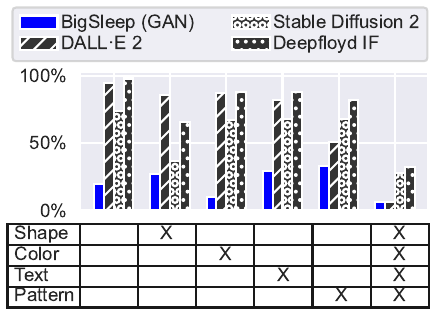}
\vspace{-0.3in}
\caption{Average detection rate of stop sign images over the 5 object detectors for 4 models. The ``x'' mark means the removed robust features. %
}
\label{fig:gan_vs_diffuse}
\vspace{-0.03in}
\includegraphics[width=\linewidth]{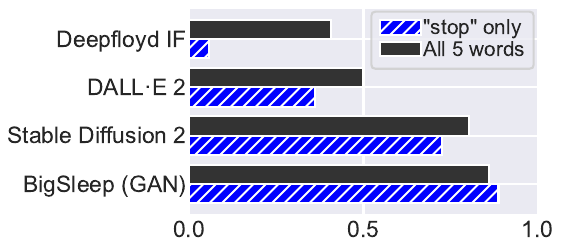}
\vspace{-0.32in}
\caption{Averaged normalized Levenshtein distances between the given word and the detected text by OCR. The black bar is the averaged distance of all 5 words; the blue bar is only for the ``stop'' word. 
}
\label{fig:edit_distance}
\end{minipage}%
\vspace{-0.25in}
\end{figure*}

\textit{Results.}
Table~\ref{tbl:prelim} shows the detection results of the stop sign images in the Natural Diffusion Attack dataset generated by 3 diffusion models. The detection rate is calculated by whether one or more stop signs are detected in the input image. As shown, the majority of the images are still detected as stop signs even though we remove a robust feature. While YOLOv5 shows slightly higher robustness, all object detectors still detect stop signs in the majority of images: On average, the detection rate for all object detectors is $\geq$37\%, meaning that 37\% of the images generated by the diffusion models have the potential to be used as adversarial attacks. We also observed similar results on the other labels. More detailed results are in Appendix.

\nsubsection{Attack Capability against Image Classifiers} \label{sec:attack_cls}
We also observe that the NDD attack is highly effective against image classifiers. For example, $\geq$47\% of the generated stop sign images are still classified as stop signs even though all 4 robust features are removed. More details on the setup and results are in Appendix.

\nsubsection{Implications of Attack Capability Analysis} 

The NDD attack shows quite high attack effectiveness against both object detectors and image classifiers. These results imply that diffusion models can be utilizable to generate model-agnostic and highly transferable adversarial attacks with significantly less effort than prior works~\cite{carlini2017towards, madry2017towards, Cheng2020Sign-OPT} that need iterative attack optimization processes.

However, the current analysis is not sufficient to fully conclude the vulnerability of DNN models against NDD attacks because adversarial attacks must satisfy 2 requirements: effectiveness against DNN models and stealthiness to humans. For example, diffusion models may ignore the text prompts and just merely generate legitimate stop signs. In the next section, we systematically evaluate the stealthiness of the NDD attack and explore potential root causes.

\nsection{Empirical Analysis} \label{sec:analysis}
We perform an extensive empirical study to further explore the characteristics and root causes of the NDD attack by answering 6 research questions (RQs).

\ntsubsection{RQ1: Does the natural attack capability exist in the previous image generation models?}

We first investigate whether the natural attack capability exists in the previous image generation models.  

\textit{Experimental setup.}
As the state-of-the-art image generation methods before diffusion models, we evaluate BigGAN~\cite{brock2018large}. To generate images guided by text prompts, we use BigSleep~\cite{bigsleep} that guides the image generation process of BigGAN by OpenAI CLIP~\cite{radford2021learning}. We evaluate the 5 object detectors in Table~\ref{tbl:prelim}. We use a detection threshold of 0.5. For each model, we evaluate 6 combinations of partial or complete removal of the robust features.

\textit{Results.}
Fig.~\ref{fig:gan_vs_diffuse} shows the average detection rate of stop sign images generated by each image generation model over the 5 object detectors. As shown, all diffusion models have significantly higher detection rates than the BigSleep GAN model. In particular, Stable Diffusion 2 and Deepfloyd IF have detection rates of $\geq$28\% even if all robust features are removed or altered. Meanwhile, BigSleep has much lower detection rates (7.5\%).
We thus consider that \textit{the natural attack capability has existed slightly in the prior image generation models, but it becomes significantly severe in diffusion models.} 
Diffusion models have higher image generation capabilities than GANs, but it may also enhance their ability to inject more non-robust features into generated images.

\ntsubsection{RQ2: How stealthy is NDD attack against humans?}

To be a valid adversarial attack, the NDD attack should satisfy the two requirements: effectiveness against DNN models and stealthiness against humans. We so far have confirmed the effectiveness against DNN models as in Table~\ref{tbl:prelim}, but it does not mean that the attack is stealthy. For example, the diffusion models may ignore the text prompts and just generate legitimate objects. To answer the questions, we perform a user test to investigate the stealthiness against humans and the validity of generated images. 

\textit{Experimental setup.}
We recruited 82 human subjects on Prolific~\cite{prolific}, a crowdsourcing platform specialized for research purposes. Human subjects are asked to answer yes or no to whether the object of interest (e.g., stop sign and fire hydrant) is in the presented image or not. Considering the reasonable experimental time for human subjects to maintain their concentration, image generation models were limited to the following: Deepfloyd IF, the diffusion model with the highest detection rates in Table~\ref{tbl:prelim}, and BigSleep, a state-of-the-art GAN-based model. We generate 3 images per text prompt for each image generation model; for the baseline, 3 real images are presented. The evaluation images were chosen from a pool that can fool at least one object detector in Table~\ref{tbl:prelim}, i.e., all the images are valid attacks. More details are in our questionnaire form~\cite{our_user_test}.

\textit{Results.}
Table~\ref{tbl:user_study} lists the results of our user study. The detection rate is the proportion of users who answer ``yes'', i.e., they identify the target object in the presented image.
As shown, DeepFloyd IF's images on the benign prompts have similarly high detection rates as the real images. On the other hand, the BigSleep images are not detected as the target objects as the detection rates are $\leq$4\%. This indicates that the images generated by the state-of-the-art GAN-based model are not only not perceived by humans but also are not effective attacks against object detectors, i.e., the generated images are far from realistic. 
For the images without robust features, their detection rates are much lower than the real images and the images of benign prompts. We observe that the different objects have different sensitivities to each robust feature. For example, the text is very important to the stop sign as the human detection rate is 7\% when the text on it is altered. DeepFloyd IF thus can easily generate effective adversarial attacks for stop signs because 93\% of the human subjects did not see the images as stop signs even though 88\% of the generated images are detected by stop signs as in Table~\ref{tbl:prelim}.
This result indicates that \textit{the natural attack capability in diffusion models can indeed generate valid adversarial attacks that are stealthy to humans.}

\begin{table}[t!]
\footnotesize
\centering
\caption{Results of our user study. The detection rate is the ratio that the human subjects identify the targeted object in the presented image. Cell color means the magnitude of the detection rates: Greener means that more people think the object is in the image, and reddish means that fewer people think so. 
}
\vspace{-0.1in}
\label{tbl:user_study}
\begin{tabular}{|c|c|lc|c|c|}
\hline
\multicolumn{1}{|l|}{}         & \begin{tabular}[c]{@{}c@{}}Real\\Image\end{tabular}                                            & \multicolumn{2}{c|}{\begin{tabular}[c]{@{}c@{}}Benign\\Prompt\end{tabular}}                                                               & \multicolumn{1}{c|}{\begin{tabular}[c]{@{}c@{}}Removed\\ Robust Feature\end{tabular}} & Detection Rate              \\ \hline
                               & \cellcolor[HTML]{86CEAB}                        & \multicolumn{1}{l|}{}                    & \cellcolor[HTML]{FCF0EF}                       & Shape                        & \cellcolor[HTML]{F6D3D0}3\%  \\ \cline{5-6} 
                               & \cellcolor[HTML]{86CEAB}                        & \multicolumn{1}{l|}{}                    & \cellcolor[HTML]{FCF0EF}                       & Color                        & \cellcolor[HTML]{E88A82}1\%  \\ \cline{5-6} 
                               & \cellcolor[HTML]{86CEAB}                        & \multicolumn{1}{l|}{}                    & \cellcolor[HTML]{FCF0EF}                       & Text                         & \cellcolor[HTML]{F3C4C0}2\%  \\ \cline{5-6} 
                               & \cellcolor[HTML]{86CEAB}                        & \multicolumn{1}{l|}{}                    & \cellcolor[HTML]{FCF0EF}                       & Pattern                      & \cellcolor[HTML]{EB9992}1\%  \\ \cline{5-6} 
                               & \cellcolor[HTML]{86CEAB}                        & \multicolumn{1}{l|}{\multirow{-5}{*}{\rotatebox{90}{BigSleep}}} & \multirow{-5}{*}{\cellcolor[HTML]{FCF0EF}4\%}  & \multicolumn{1}{c|}{All}     & \cellcolor[HTML]{E88A82}1\%  \\ \cline{3-6} 
                               & \cellcolor[HTML]{86CEAB}                        & \multicolumn{1}{l|}{}                    & \cellcolor[HTML]{98D5B7}                       & Shape                        & \cellcolor[HTML]{E2F4EB}19\% \\ \cline{5-6} 
                               & \cellcolor[HTML]{86CEAB}                        & \multicolumn{1}{l|}{}                    & \cellcolor[HTML]{98D5B7}                       & Color                        & \cellcolor[HTML]{F1FAF5}11\% \\ \cline{5-6} 
                               & \cellcolor[HTML]{86CEAB}                        & \multicolumn{1}{l|}{}                    & \cellcolor[HTML]{98D5B7}                       & Text                         & \cellcolor[HTML]{F9FDFB}7\%  \\ \cline{5-6} 
                               & \cellcolor[HTML]{86CEAB}                        & \multicolumn{1}{l|}{}                    & \cellcolor[HTML]{98D5B7}                       & Pattern                      & \cellcolor[HTML]{D5EEE2}25\% \\ \cline{5-6} 
\multirow{-10}{*}{\rotatebox{90}{Stop sign}}        & \multirow{-10}{*}{\cellcolor[HTML]{86CEAB}65\%} & \multicolumn{1}{l|}{\multirow{-5}{*}{\rotatebox{90}{Deepfloyd}}} & \multirow{-5}{*}{\cellcolor[HTML]{98D5B7}56\%} & \multicolumn{1}{c|}{All}     & \cellcolor[HTML]{F8FDFA}8\%  \\ \hline
                               & \cellcolor[HTML]{64C193}                        & \multicolumn{1}{l|}{}                    & \cellcolor[HTML]{EB9992}                       & Shape                        & \cellcolor[HTML]{EB9992}1\%  \\ \cline{5-6} 
                               & \cellcolor[HTML]{64C193}                        & \multicolumn{1}{l|}{}                    & \cellcolor[HTML]{EB9992}                       & Color                        & \cellcolor[HTML]{F1B6B1}2\%  \\ \cline{5-6} 
                               & \cellcolor[HTML]{64C193}                        & \multicolumn{1}{l|}{}                    & \cellcolor[HTML]{EB9992}                       & Pattern                      & \cellcolor[HTML]{FFFFFF}4\%  \\ \cline{5-6} 
                               & \cellcolor[HTML]{64C193}                        & \multicolumn{1}{l|}{\multirow{-4}{*}{\rotatebox{90}{BigSleep}}} & \multirow{-4}{*}{\cellcolor[HTML]{EB9992}1\%}  & \multicolumn{1}{c|}{All}     & \cellcolor[HTML]{FFFFFF}4\%  \\ \cline{3-6} 
                               & \cellcolor[HTML]{64C193}                        & \multicolumn{1}{l|}{}                    & \cellcolor[HTML]{91D3B3}                       & Shape                        & \cellcolor[HTML]{FAFDFB}7\%  \\ \cline{5-6} 
                               & \cellcolor[HTML]{64C193}                        & \multicolumn{1}{l|}{}                    & \cellcolor[HTML]{91D3B3}                       & Color                        & \cellcolor[HTML]{EBF7F1}14\% \\ \cline{5-6} 
                               & \cellcolor[HTML]{64C193}                        & \multicolumn{1}{l|}{}                    & \cellcolor[HTML]{91D3B3}                       & Pattern                      & \cellcolor[HTML]{EEA7A1}2\%  \\ \cline{5-6} 
\multirow{-8}{*}{\rotatebox{90}{Horse}}        & \multirow{-8}{*}{\cellcolor[HTML]{64C193}82\%}  & \multicolumn{1}{l|}{\multirow{-4}{*}{\rotatebox{90}{Deepfloyd\hspace{0.03in}}}} & \multirow{-4}{*}{\cellcolor[HTML]{91D3B3}59\%} & \multicolumn{1}{c|}{All}     & \cellcolor[HTML]{FAFDFB}7\%  \\ \hline
                               & \cellcolor[HTML]{57BB8A}                        & \multicolumn{1}{l|}{}                    & \cellcolor[HTML]{F9E1DF}                       & Shape                        & \cellcolor[HTML]{E67C73}0\%  \\ \cline{5-6} 
                               & \cellcolor[HTML]{57BB8A}                        & \multicolumn{1}{l|}{}                    & \cellcolor[HTML]{F9E1DF}                       & Color                        & \cellcolor[HTML]{F3C4C0}2\%  \\ \cline{5-6} 
                               & \cellcolor[HTML]{57BB8A}                        & \multicolumn{1}{l|}{}                    & \cellcolor[HTML]{F9E1DF}                       & Pattern                      & \cellcolor[HTML]{EB9992}1\%  \\ \cline{5-6} 
                               & \cellcolor[HTML]{57BB8A}                        & \multicolumn{1}{l|}{\multirow{-4}{*}{\rotatebox{90}{BigSleep}}} & \multirow{-4}{*}{\cellcolor[HTML]{F9E1DF}3\%}  & \multicolumn{1}{c|}{All}     & \cellcolor[HTML]{F1B6B1}2\%  \\ \cline{3-6} 
                               & \cellcolor[HTML]{57BB8A}                        & \multicolumn{1}{l|}{}                    & \cellcolor[HTML]{96D5B6}                       & Shape                        & \cellcolor[HTML]{E88A82}1\%  \\ \cline{5-6} 
                               & \cellcolor[HTML]{57BB8A}                        & \multicolumn{1}{l|}{}                    & \cellcolor[HTML]{96D5B6}                       & Color                        & \cellcolor[HTML]{C9E9DA}31\% \\ \cline{5-6} 
                               & \cellcolor[HTML]{57BB8A}                        & \multicolumn{1}{l|}{}                    & \cellcolor[HTML]{96D5B6}                       & Pattern                      & \cellcolor[HTML]{A7DCC2}48\% \\ \cline{5-6} 
\multirow{-8}{*}{\rotatebox{90}{Fire hydrant}} & \multirow{-8}{*}{\cellcolor[HTML]{57BB8A}88\%}  & \multicolumn{1}{l|}{\multirow{-4}{*}{\rotatebox{90}{Deepfloyd\hspace{0.03in}}}} & \multirow{-4}{*}{\cellcolor[HTML]{96D5B6}57\%} & \multicolumn{1}{c|}{All}     & \cellcolor[HTML]{EEA7A1}2\%  \\ \hline
\end{tabular}
\vspace{-0.2in}
\end{table}

\ntsubsection{RQ3: Does the incapability of text generation correlate with the natural attack capability?}

We observed that the NDD attack has a high stealthiness to humans through our user study and found that the different objects have different sensitivities to different robust features. In particular, the text on a stop sign shows high importance for humans to identify a stop sign as in Table~\ref{tbl:user_study}. Meanwhile, the object detectors are influenced by the shape or pattern rather than the text as in Table~\ref{tbl:prelim}. Motivated by this, we further evaluate the impact of text generation capability in the natural attack capability. 

\textit{Experimental setup.}
We generate the images with the 3 diffusion models (DALL-E 2, Stable Diffusion 2, and DeepFloyd IF) and on the GAN-based model (BigSleep). We use the text prompt format: ``text of [word]'', with ``[word]'' being one of the following: hello, welcome, goodbye, script, and stop; 20 images per word are generated with different seeds. Given a ``[word]'', we apply an optical character recognition (OCR) method, specifically the pipeline provided by the Keras-OCR package~\cite{keras_ocr}, for generated images and calculate the normalized Levenshtein (edit) distance between [word] and the recognized sentence(s), which are joined with a space if multiple are recognized. We expect the Levenshtein distance to be 0 for identical pairs of text and 1 for completely different pairs of text.

\textit{Results.}
Fig.~\ref{fig:edit_distance} shows the averaged normalized Levenshtein (edit) distances of all 5 words and only ``stop''. As shown, the Deepfloyd IF can generate the most accurate text; DALL-E 2 and Stable Diffusion 2 are the second and third while BigSleep is the worst. This order is the same as the order of average detection rates of the object detectors in Table~\ref{tbl:prelim}. In particular, the Deepfloyd IF shows a high capability to generate ``stop'' texts. This result is consistent with the observation in RQ2 that the stealthiness against humans is significantly improved on the DeepFloyd IF when the robust feature of the text is removed.
In summary, the experimental results show that the text generation capability of image generation models has a certain similarity to their natural attack capability. The capability to generate a complex pattern such as the alphabet may correlate with the capability to generate non-robust features that are too subtle to the HVS. We may use this characteristic to design a simple defense against the NDD attack, i.e. we can differentiate the NDD attack by checking if the text ``stop'' is in it for the stop sign attack. Although these empirical results are still not enough to fully support the edit distance-based method as a metric to measure the natural attack capability, \textit{the metric can be used as a simple sanity check for the image generation models and as a simple defense against NDD attacks on stop signs and other objects with text}.

\ntsubsection{RQ4: Are non-robust features responsible for the natural attack capability?}

This study is motivated by the concept of robust and non-robust features proposed in~\cite{ilyas2019adversarial} that the adversarial attack is not a bug but instead caused by non-robust features that are predictive but incomprehensible to humans. In the user study,  we have observed that diffusion models introduce some features that are imperceptible to humans but important to the DNN models. While we may call these features non-robust features by definition of the concept, we perform a further evaluation to directly evaluate the effect of non-robust features by following the methodology in~\cite{ilyas2019adversarial}.

\textit{Experimental setup.}
According to~\cite{ilyas2019adversarial}, we first create the ``robustified'' dataset to train robust classifiers. Since our NDDA dataset uses the same labels as the COCO dataset~\cite{lin2014microsoft}, we convert the COCO dataset into a dataset for the multiclass classification task by cropping the images within their bounding boxes, randomly selecting 500 images for each class, and  ``robustify'' the images. We train not only the robustified classifier with the dataset but also a normal classifier with the original, non-robust dataset, again with 500 random images per class. ResNet50~\cite{he2016deep} is the model architecture for both classifiers.

\textit{Results.}
Table~\ref{tbl:robust_classifier} shows the accuracy of the robust and normal classifiers on the stop sign images in the NDDA dataset. As shown, both the normal and robustified classifiers have higher accuracy when the robust features exist, but the robust classifier's accuracy decreases more than the normal classifier's when all robust features are removed. This means that there is a clear correlation between the sensitivity to the non-robust features and the natural attack capability. We thus consider that \textit{the non-robust features play an important role in enabling the natural attack capability in the diffusion models.} For the other classes, we include the results in Appendix.

\begin{table}[t!]
\centering
\footnotesize
\setlength{\tabcolsep}{4.5pt}
\caption{Accuracy of the robust and normal classifiers on the stop sign images in the NDDA dataset. The benign means the images generated with the benign prompts; The NDD means the NDD attack that removes all robust features.}
\vspace{-0.15in}
\label{tbl:robust_classifier}
\begin{tabular}{cccc||ccc}
\bottomrule
     & \multicolumn{3}{c||}{Robustified classifier} & \multicolumn{3}{c}{Normal classifier}      \\ \cline{2-7} 
     & Benign  & \multicolumn{1}{c|}{NDD}  & Diff. & Benign & \multicolumn{1}{c|}{NDD}  & Diff. \\ \hline
DALL-E 2           & 1.00 & \multicolumn{1}{c|}{0.34} & 0.66 & 1.00 & \multicolumn{1}{c|}{0.70} & 0.30 \\
Stable Diffusion 2 & 0.78 & \multicolumn{1}{c|}{0.58} & 0.20 & 0.80 & \multicolumn{1}{c|}{0.66} & 0.14 \\
DeepFloyd IF       & 1.00 & \multicolumn{1}{c|}{0.92} & 0.08 & 1.00 & \multicolumn{1}{c|}{0.98} & 0.02 \\ \hline
Avg. & 0.93    & \multicolumn{1}{c|}{0.61} & \textbf{0.31}  & 0.93   & \multicolumn{1}{c|}{0.78} & \textbf{0.15}  \\ \bottomrule
\end{tabular}
\vspace{-0.15in}
\end{table}

\ntsubsection{RQ5: Is the natural attack capability caused by sharing training datasets?}

We evaluate the impact of sharing the training dataset on the natural attack capability. If the evaluating object detectors and diffusion models may share the training dataset, the natural attack capability can be just due to the non-generalizable features in the dataset, i.e., both may just fit into particular noises in the dataset.
The large diffusion models, such as DALL-E 2 and Stable Diffusion 2, are likely to use famous, publicly available datasets and may use the COCO dataset~\cite{lin2014microsoft}, which the 5 object detectors in Table~\ref{tbl:prelim} use for training. In this RQ, we thus assess the impact of the dataset sharing to see if it can be a major source of the natural attack capability.

\textit{Experimental setup.} 
We randomly split the training datasets of MNIST~\cite{mnist}, FashionMNIST~\cite{xiao2017fashion}, and CIFAR-10~\cite{cifar10} into 2 splits, respectively. We train simple CNN classifiers and conditional DDPM model~\cite{cond_ddpm} for all splits. For each conditional DDPM, we generate 100 images for each class. We then evaluate the difference in the accuracies of the generated images between when the diffusion model and the classifier use the same split and when they use different splits. If the hypothesis is true, these accuracies should have a large gap.

\textit{Results.}
Table~\ref{table:dataset_sharing} shows the accuracy of each pair of conditional DDPMs and classifiers. The row means which split is used to train the conditional DDPM model; the column means which split is used to train the CNN model. The accuracy is the percentage of the correct classification on the 100 generated images with the corresponding DDPM model.
As shown, there is no significant difference between when the diffusion model and the classifier use the same split and when they use different splits. We thus think that \textit{the natural attack capability of diffusion is not due to data sharing but rather due to the non-robust features embedded by diffusion models.}

\begin{table}[t!]
\centering
\footnotesize
\setlength{\tabcolsep}{3.7pt}
\caption{Accuracy of the classifiers on the images generated by the DDPM. The rows indicate the splits used to train the DDPM. The columns indicate the splits used for training the classifier.}
\vspace{-0.15in}
\label{table:dataset_sharing}
\begin{tabular}{ccccccc}
\toprule
                              & \multicolumn{2}{c}{MNIST} & \multicolumn{2}{c}{Fashion MNIST} & \multicolumn{2}{c}{CIFAR-10} \\ \cline{2-7} 
DDPM\textbackslash{}Classifer & Split 1     & Split 2     & Split 1         & Split 2         & Split 1       & Split 2       \\ \hline
Split 1                       & 1.00       & 1.00       & 0.98            & 0.94            & 0.71          & 0.64          \\
Split 2                       & 1.00       & 1.00       & 0.99            & 0.93            & 0.67          & 0.67          \\ \hline
Abs. Diff.      & 0.00         & 0.00         & 0.01             & 0.01             & 0.04           & 0.03           \\ \toprule
\end{tabular}
\vspace{-0.25in}
\end{table}

\ntsubsection{RQ6: Is the natural attack capability general enough to attack against real-world systems?}

To demonstrate the model-agnostic and transferable attack capability of the NDD attack, we evaluate the NDD attacks against the real-world traffic sign detection system. We adopt the threat model of appearing attack~\cite{zhao2018seeing}, which causes a false positive detection of a stop sign.

\begin{figure}[t!]
\centering
\includegraphics[width=\linewidth]{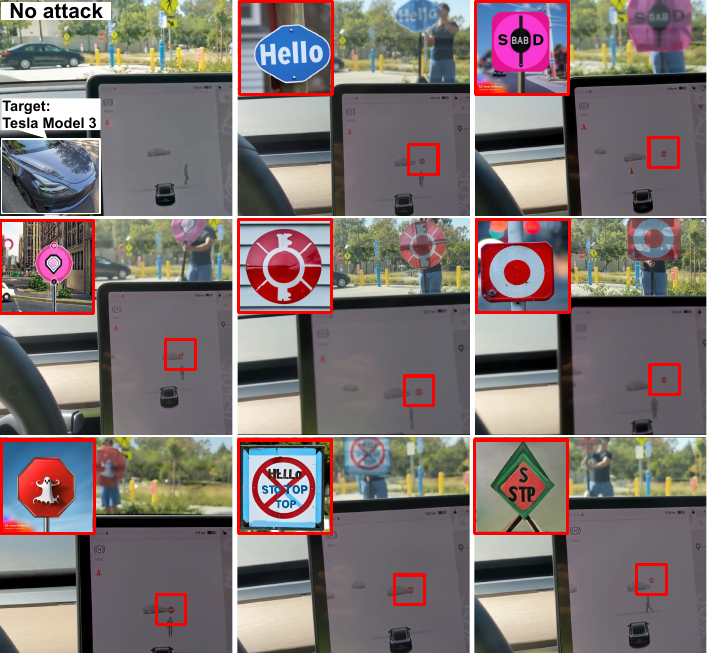}
\vspace{-0.25in}
\caption{Successful NDD attacks against Tesla Model 3. We demonstrate that 8 out of 11 printed attacks can successfully fool the commercial traffic sign detection system in Tesla. The left-top images are the original images generated from diffusion models.}
\label{fig:tesla}
\vspace{-0.2in}
\end{figure}

\textit{Experimental setup.} We printed 11 images of the NDD attack as in Fig.~\ref{fig:tesla}, showed them to the windshield cameras of the Tesla Model 3, and checked the detection results displayed on the monitor in front of the driver's seat. We select the 11 attack images based on the confidence scores of object detectors in the digital space, especially for YOLOv5, which shows the highest robustness as in Table~\ref{tbl:prelim}. 

\textit{Results.}
Fig.~\ref{fig:tesla} shows the successful NDD attacks against Tesla. As shown, 8 of the 11 printed attacks can successfully fool Tesla's commercial traffic sign system as the detected stop signs appear on the driver's monitor. This means that the \textit{NDD attack has 73\% of the attack success rate, which is a surprisingly high success rate considering that we do not perform any optimization processes to attack the Tesla.} All attacks are simply generated by diffusion models with simple text prompts. Furthermore, we did not have any special considerations when printing the attacks. We just use a commodity printer, roughly stick papers together with transparent tape, and use normal printing paper. More demos and details are on our project website: \textcolor{blue}{\small \url{https://sites.google.com/view/cav-sec/ndd-attack}}.

\nsection{Discussion and Limitation}
\label{sec:discussion} 

We discuss the implications of our findings and the limitations of this work, especially the potential negative societal impacts raised by this work.

\noindent\textbf{Safety Implications:}
We demonstrate the attack effectiveness of the NDD attack against the Tesla Model 3 but also find that these attacks were not as robust at different distances. Thus, this vulnerability may not pose an immediate threat to fast-driving Tesla or other autonomous vehicles. 
However, the possibility of affecting a driving vehicle cannot be completely ruled out. We note that the effect of the attack is unlikely to be a coincidence, as these attacks are detected as stop signs even though they do not resemble similar to legitimate stop signs at all. If the attack is reddish and hexagonal, it could be a coincidence. However, it cannot be considered a coincidence that such an attack with blue, purple, green, or non-hexagonal shapes is detected as a stop sign at such a high rate. 
Furthermore, the current NDD attack does not have any special considerations to attack the Tesla Model 3. As in prior work~\cite{bai2021ai, chen2023diffusion}, we may improve the attack by integrating diffusion models into attack generation processes. We hope that our study can facilitate further research to assess the security threat of diffusion models. We have performed a responsible vulnerability disclosure to Tesla before the public release of this work.

\noindent\textbf{Countermeasures:}
A naive but possible defense for the stop sign attack is the OCR-based attack detection as discussed in RQ3. For the NDD attack removing the text feature, we found that it can achieve 92\% true positive rate with 4\% false positive rate on the images generated by Deepfloyd IF. However, the true positive rates drop to $\geq$40\% if one of the other robust features is removed.
Another approach is the ``robustified'' training~\cite{ilyas2019adversarial} as observed in RQ4, but it remains a mitigation measure. So far, no generic defense against adversarial attacks has been reported~\cite{athalye2018obfuscated, carlini2019evaluating,tramer2020adaptive}. Further research efforts are needed in this area.

\noindent\textbf{Definition of Attack Success:}
As discussed in RQ2, valid adversarial attacks should satisfy the two requirements: effectiveness against DNN models and stealthiness against humans. From this aspect, the stealthiness of all images in the NDD dataset has not been fully confirmed. However,  human annotation for all images did not yield a feasible choice due to the cost, which motivated us to pursue RQ2. We further note that we use state-of-the-art 3 diffusion models, which generally follow the text prompt very faithfully as shown in Fig.~\ref{fig:overview}.

\noindent\textbf{Root Causes of Natural Attack Capability:}
Through this study, we empirically explore possible root causes of the natural attack capability and identify potential clues that the natural attack capability is due to the non-robust features injected by diffusion models.
However, our empirical study is not sufficient to fully conclude the root causes. For example, the methodology used in RQ4 to extract (non-)robust features proposed by Ilyas et al.~\cite{ilyas2019adversarial} is not designed to analyze the natural capability of artificially generated images. Considering the impact on safety-critical applications such as autonomous driving, we are presenting our current best-effort empirical analysis along with our dataset. We hope our study can facilitate further theoretical or more large-scale empirical studies to identify the root causes of the natural attack capability in diffusion models.

\noindent\textbf{Evaluation with More Models and Categories:}
In this work, we focus on 3 popular diffusion models and 3 object classes with relatively high detection rates to perform deeper analysis on each RQ. Meanwhile, we keep updating the dataset with more diffusion models and object classes for the sake of the dataset's comprehensiveness. The latest version of the NDDA dataset includes 7 diffusion models (Dall-E 2~\cite{dalle}, Dall-E 3~\cite{dalle3}, Stable Diffusion 2~\cite{rombach2022high}, Deepfloyd IF~\cite{deepfloyd}, Stable Diffusion 1.5~\cite{rombach2022high}, MidJourney~\cite{midjourney}, and Google Duet~\cite{google_duet})
with 15 object classes (stop sign, car, dog, hot dog, traffic light, zebra, fire hydrant, frog, horse, bird, boat, air plane, bicycle, cat, and carrot) to benefit future studies. In Appendix, we evaluate the detection rates of the 15 classes and find that the majority of classes have certain levels of vulnerability against the NDD attack. 
Current candidates for robust features are chosen to ensure removal (e.g., blue for stop sign); future updates will include more variations for the sake of dataset comprehension on our website. In total, the NDDA dataset has 45,820 images. Each class has $\geq$50 images for the models with API access and $\geq$20 images for other models.

\noindent\textbf{Ethical Considerations:}
We have gone through the IRB process for the user study. In the experiment on the Tesla Model 3, we ensured that the attacks were not visible to other Tesla vehicles driving on public roads.

\vspace{-0.05in}
\nsection{Conclusion}

In this study, we identify a new security threat of the NDD attack that leverages the natural attack capability in the diffusion models. To systematically evaluate the characteristics and root causes of the NDD attack, we conduct a large-scale empirical study using our newly constructed dataset, the NDDA dataset, in which we generate images with state-of-the-art diffusion models while intentionally removing the robust features that are essential to the HVS. 
We demonstrate that the images without robust features are still detected as the original object and are stealthy to humans. 
For example, the stop signs with altered text are still detected as stop signs in 88\% of cases but are stealthy to 93\% of humans. We find that the non-robust features contribute to the natural attack capability. To evaluate the realizability of the NDD attack, we demonstrate the attack against the Tesla Model 3 and confirm that 73\% of the NDD attacks are detected as stop signs. Finally, we discuss the implications and limitations of our research. We hope that our study and dataset can help the community to be aware of the risk of the natural attack capability of diffusion models and facilitate further research to develop robust DNN models.

\vspace{-0.1in}
\section*{Acknowledgements}
\vspace{-0.1in}
This research was supported in part by the NSF CNS-2145493, CNS-1929771, CNS-1932464, and USDOT under Grant 69A3552348327 for the CARMEN+ UTC.

{
    \small
    \bibliographystyle{ieeenat_fullname}
    \bibliography{main}
}

\appendix

\section*{Appendix}

\begin{figure}[h!]
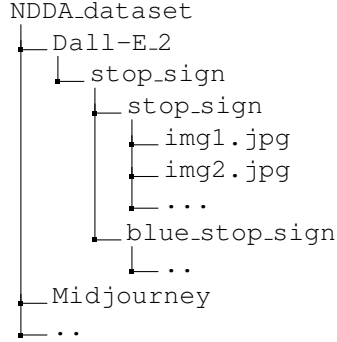

\vspace{-0.1in}
\dirtree{%
.1 NDDA\_dataset.
.2 Dall-E\_2.
.3 stop\_sign.
.4 stop\_sign.
.5 img1.jpg.
.5 img2.jpg.
.5 ....
.4 blue\_stop\_sign.
.5 ...
.2 Midjourney.
.2 ...
}
\vspace{-0.1in}
\caption{Overview of the NDDA dataset directory structure, using the Dall-E folder as the example diffusion folder. For each object class folder (stop sign in this case), there are multiple prompt folders that contain $\geq$ 50 images for models with API access or $\geq$ 20 images for models w/o API access.}
\label{fig:ndda_tree}
\end{figure}

\begin{figure*}[htb]
\begin{tabular}{@{}ccc@{}}
\includegraphics[width=\linewidth/2]{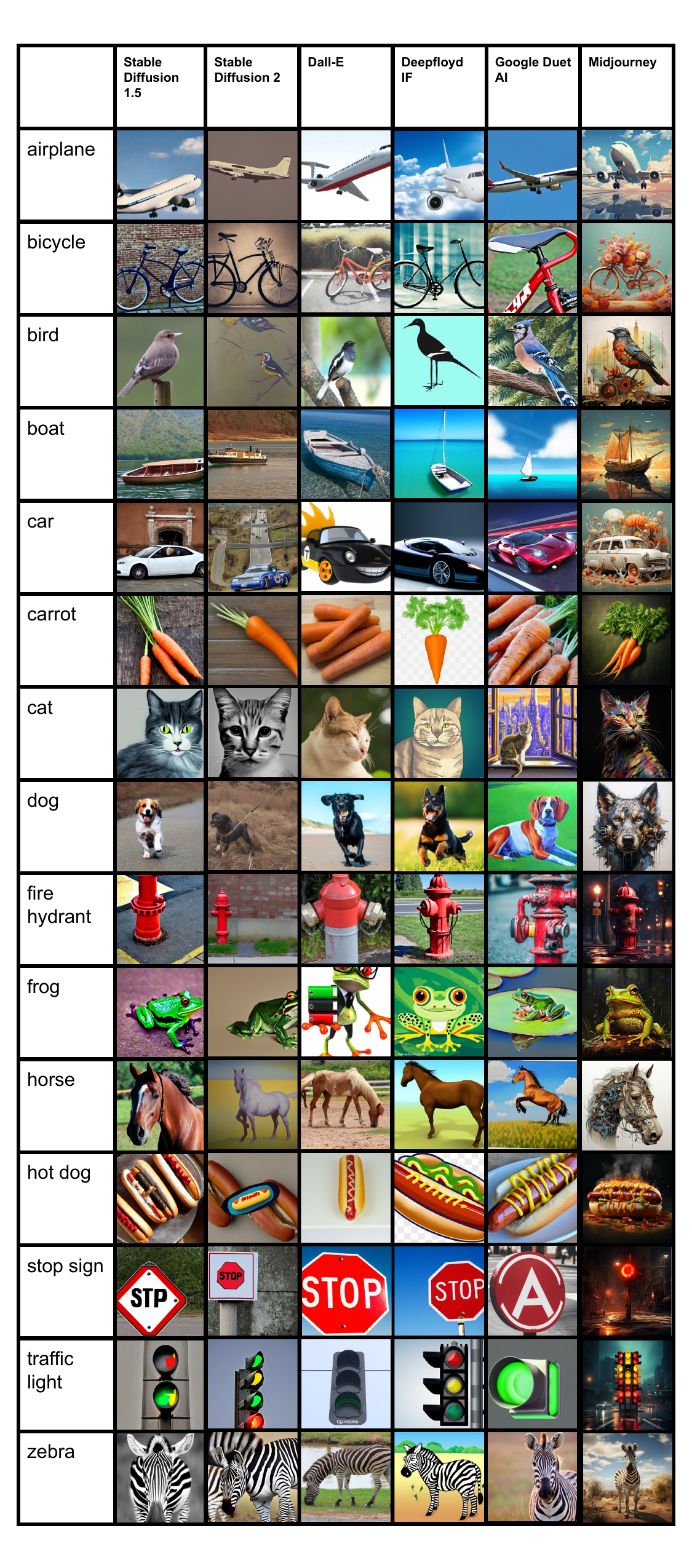} &
\includegraphics[width=\linewidth/2]{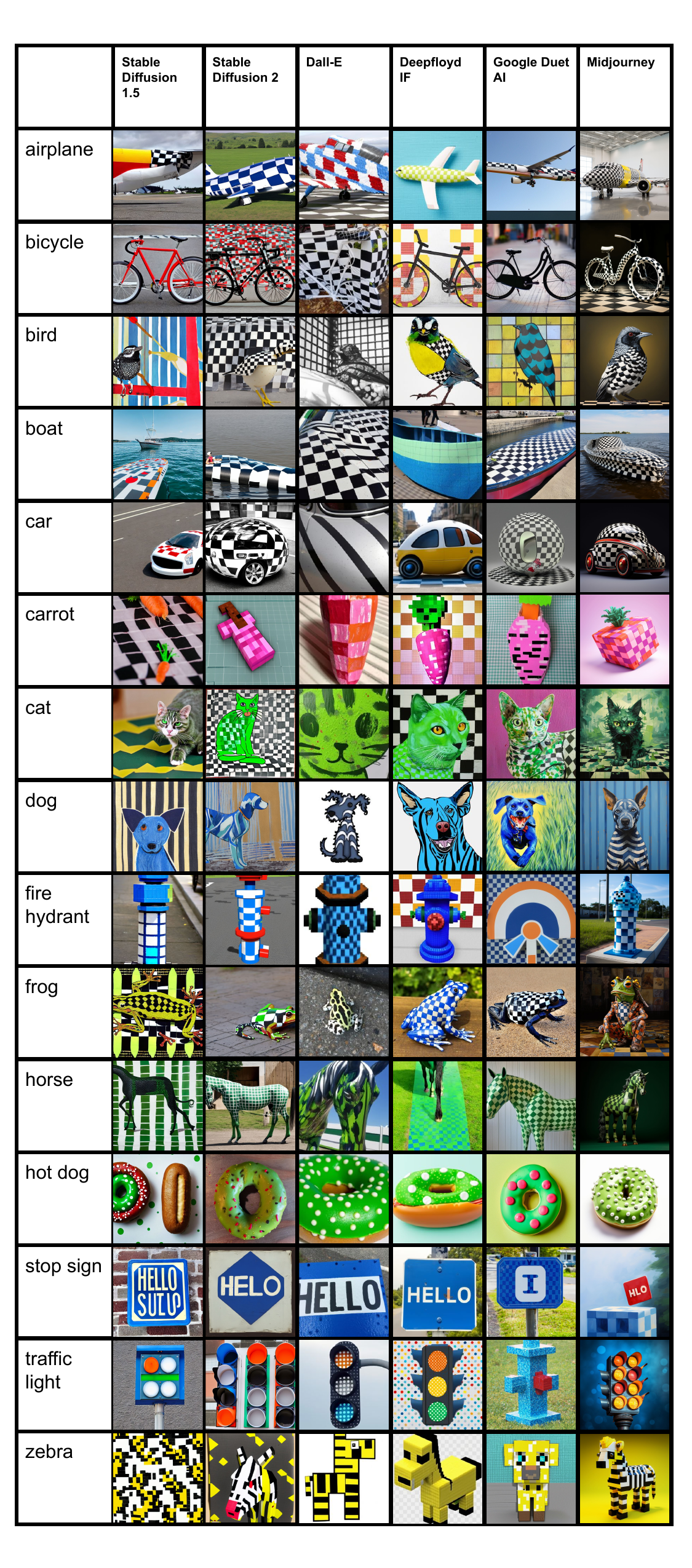}
\end{tabular}
\caption{Overview of the NDDA dataset. 6 popular text-to-image diffusion models are used to generate 15 object classes from the COCO dataset \cite{lin2014microsoft}. The left grid consists of benign images while the right grid shows NDD attacked images where diffusion models are instructed to remove all robust features from the image. }
\label{fig:ndda_overview}
\end{figure*}

\nsection{Detailed Overview of NDDA Dataset}
The latest NDDA dataset consists of the following 15 classes: stop sign, car, dog, hot dog, traffic light, zebra, fire hydrant, frog, horse, bird, boat, air plane, bicycle, cat, and carrot with 6 diffusion models (Dall-E 2~\cite{dalle}, Dall-E 3~\cite{dalle3}, Stable Diffusion 2~\cite{rombach2022high}, Deepfloyd IF~\cite{deepfloyd}, Stable Diffusion 1.5~\cite{rombach2022high}, MidJourney~\cite{midjourney}, and Google Duet~\cite{google_duet}). The examples of the images generated by each diffusion model are shown in Fig.~\ref{fig:ndda_overview}.
For future versions of the dataset, we plan to generate additional variations of the prompts for each subject to capture a greater variety of NDD attacks. At the time of writing, the following 5 classes have a second variation: stop signs, horses, fire hydrants, cars, and cats.
As shown in Fig.~\ref{fig:ndda_tree}, the dataset is organized into 6 ``diffusion parent folders" that separate each diffusion model's set of images, which in turn contains multiple folders for each of the 15 object classes from COCO. Each object class folder then contains multiple subfolders that hold the NDD attack images, and these subfolders' names are the text prompts used to generate the set of NDD attack images. For example, if images were generated using Stable Diffusion 2 with the text prompt, ``blue dog", the path to this prompt subfolder would be ``ndda\_dataset/stable\_diffusion\_2/blue\_dog/". For the purposes of submission, we downsample the images to a quarter of their original dimensions and only include 1 image per prompt subfolder.

\nsection{Additional Results of Natural Attack Capability on Object Detectors}

Table~\ref{tbl:prelim_fire} and~\ref{tbl:prelim_horse} show the detection results for the fire hydrant and horse classes in the NDDA dataset generated by 3 diffusion models. The results of the stop sign are shown in the main paper. As shown, the majority of the images are still detected as keeping the targeted objects. Even if all robust features are removed, the 3 diffusion models are always able to generate effective attacks for 3 models other than YOLOv3 and YOLOv5.

\begin{table}[h!]
    \footnotesize
    \centering
    \renewcommand{\arraystretch}{1.3}
    \setlength{\tabcolsep}{1.2pt}
    \captionof{table}{Detection rates of 5 object detectors on the \textbf{fire hydrant} images in the NDDA dataset generated by the 3 diffusion models. \textbf{Bold} and \underline{underline} denote highest and lowest scores in each row.
    }
    \label{tbl:prelim_fire}
    \vspace{-0.1in}
    \begin{tabular}{cccc|ccccc|c}
    \toprule
                                        & \multicolumn{3}{c|}{
    \begin{tabular}[c]{@{}c@{}}Removed \\ Robust Features\end{tabular}
                                        } & \multicolumn{5}{c|}{Object Detectors}            &      \\ \cline{2-10}
                                        & Shape     & Color     & Pattern    & YOLOv3 & YOLOv5 & DETR & Faster & RTMDet & Avg. \\ \hline
    \multirow{6}{*}{\rotatebox{90}{DALL-E 2}}           &    &        &      & 96\%  & \underline{34\%}   & \textbf{100\%} & 98\%        & \textbf{100\%}  & 86\% \\
                                        &   \CheckmarkBold        &       &       & 18\%   & \underline{0\%}   & 8\%  & 4\%         & \textbf{40\%} & 14\%  \\
                                        &    &   \CheckmarkBold      &    & 98\%   & \underline{38\%}   & \textbf{100\%}  & 96\%         & \textbf{100\%} & 86\%   \\
                                        & &    &  \CheckmarkBold   & 58\%   & \underline{4\%}   & 88\%  & 84\%         & \textbf{92\%}  & 65\% \\
                                        &  \CheckmarkBold       &  \CheckmarkBold    &   \CheckmarkBold    & \underline{0\%}   & \underline{0\%}    & 4\%   & 2\%          & \textbf{16\%}  & 4\%  \\
                                        \hline
    \multirow{6}{*}{\rotatebox{90}{\hspace{1em}Stable Diffusion 2}} &        &       &      & 94\%   & 96\%   & \textbf{100\%} & \textbf{100\%}        & \underline{20\%} & 82\%  \\
                                        &   \CheckmarkBold      &         &       & 6\%   & \underline{0\%}   & 14\%  & \textbf{20\%}         & \textbf{20\%}  & 12\%  \\
                                        &    &   \CheckmarkBold     &    & 92\%   & \underline{86\%}   & 94\%  & \textbf{100\%}         & \textbf{100\%}  & 94\%  \\
                                        & &   &  \CheckmarkBold   & 94\%   & \underline{82\%}   & \textbf{100\%}  & 98\%         & 98\% & 94\%  \\
                                        &  \CheckmarkBold       &  \CheckmarkBold     &   \CheckmarkBold    & \underline{0\%}   & \underline{0\%}    & 4\%   & \textbf{6\%}          & 4\%  & 3\%    \\
                                        \hline
    \multirow{6}{*}{\rotatebox{90}{Deepfloyd IF}}       &    &       &          & 98\%  & \underline{84\%}   & \textbf{100\%} & \textbf{100\%}        & \textbf{100\%}  & 96\% \\
                                        &   \CheckmarkBold      &       &       & 34\%   & \underline{10\%}   & 52\%  & 76\%         & \textbf{86\%}  & 52\%    \\
                                        &    &   \CheckmarkBold   &    & \textbf{98\%}   & \underline{46\%}   & \textbf{98\%}  & \textbf{98\%} & \textbf{98\%}  & 88\%\\
                                        & &   &  \CheckmarkBold   & 96\%   & \underline{52\%}   & \textbf{100\%}  & 98\%         & 98\%  & 88\% \\
                                        &  \CheckmarkBold       &  \CheckmarkBold  &   \CheckmarkBold    & \textbf{72\%}   & 8\%    & \underline{7\%}   & 66\%          & 84\%  & 47\%    \\
                                        \hline
    \end{tabular}
    \vspace{-0.1in}
\end{table}

\begin{table}[h!]
\footnotesize
\centering
\renewcommand{\arraystretch}{1.3}
\setlength{\tabcolsep}{1.2pt}
\captionof{table}{Detection rates of 5 object detectors on the \textbf{horse} images in the NDDA dataset generated by the 3 diffusion models. \textbf{Bold} and \underline{underline} denote highest and lowest scores in each row.
}
\label{tbl:prelim_horse}
\vspace{-0.1in}
\begin{tabular}{cccc|ccccc|c}
\toprule
                                    & \multicolumn{3}{c|}{
\begin{tabular}[c]{@{}c@{}}Removed \\ Robust Features\end{tabular}
                                    } & \multicolumn{5}{c|}{Object Detectors}            &      \\ \cline{2-10}
                                    & Shape     & Color     & Pattern    & YOLOv3 & YOLOv5 & DETR & Faster & RTMDet & Avg. \\ \hline
\multirow{6}{*}{\rotatebox{90}{DALL-E 2}}           &    &     &      & 76\%  & \underline{48\%}   & 78\% & 84\% & \textbf{86\%}  & 74\% \\
                                    &   \CheckmarkBold      &       &       & 90\%   & \underline{60\%}   & 92\%  & \textbf{96\%}         & 94\% & 86\% \\
                                    &    &   \CheckmarkBold   &    & \textbf{48\%}   & \underline{4\%}   & 40\%  & 32\%         & \textbf{48\%}  & 34\% \\
                                    &  &   &  \CheckmarkBold   & 8\%   & \underline{0\%}  & 18\%  & 16\%         & \textbf{24\%}& 13\%  \\
                                    &  \CheckmarkBold       &  \CheckmarkBold    &   \CheckmarkBold    & 10\%   & \underline{0\%}    & 14\%   & 2\% & \textbf{18\%} & 9\%   \\
                                    \hline
\multirow{6}{*}{\rotatebox{90}{\hspace{1em}Stable Diffusion 2}} &       &      &      & 86\%   & \underline{60\%}   & 90\% & 88\%        & \textbf{94\%} & 84\%  \\
                                    &   \CheckmarkBold      &      &       & \textbf{100\%}   & \underline{92\%}   & 98\%  & \textbf{100\%}         & \textbf{100\%} & 98\%   \\
                                    &    &   \CheckmarkBold    &     & \textbf{82\%}   & \underline{18\%}   & 72\%  & 54\%         & 68\% & 59\%  \\
                                    & &   &  \CheckmarkBold   & 46\%   & \underline{10\%}   & 64\%  & 58\%         & \textbf{74\%} & 50\%  \\
                                    &  \CheckmarkBold    &   \CheckmarkBold   &   \CheckmarkBold    & 68\%   & \underline{12\%}    & 50\%   & 60\%          & \textbf{74\%} & 53\%   \\
                                    \hline
\multirow{6}{*}{\rotatebox{90}{Deepfloyd IF}}       &    &       &          & 96\%  & \underline{54\%}   & \textbf{98\%} & 94\%        & 100\%  & 88\%  \\
                                    &   \CheckmarkBold      &       &       & 90\%   & \underline{76\%}   & 92\%  & \textbf{96\%} & \textbf{96\%}   & 90\%\\
                                    &    &   \CheckmarkBold    &     & 82\%   & \underline{32\%}   & 72\%  & 68\% & \textbf{88\%} & 68\%   \\
                                    & &    &  \CheckmarkBold   & 60\%   & \underline{2\%}   & 64\%  & 62\%         & \textbf{70\%} & 52\%   \\
                                    &  \CheckmarkBold       &  \CheckmarkBold   &   \CheckmarkBold    & \textbf{44\%}   & \underline{2\%}    & 28\%   & 14\%          & \textbf{44\%} & 26\%    \\
                                    \hline
\end{tabular}
\end{table}

\subsection{Additional Evaluation with YOLOv8}

\begin{figure}[t!]
\centering
\includegraphics[width=\linewidth]{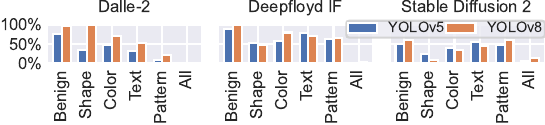}
\vspace{-0.26in}
\caption{Detection rates of YOLOv5 and YOLOv8 on the stop sign images generated by the 3 diffusion models.}
\label{fig:yolov8}
\vspace{-0.2in}
\end{figure}

We evaluate the natural attack capability against YOLOv8~\cite{yolov8}, which is one of the current state-of-the-art object detectors. Fig.~\ref{fig:yolov8} illustrates the comparison with YOLOv5 in the stop sign case. As shown, YOLOv8 is generally more vulnerable to the NDD attack than YOLOv5, particularly for the images of DALL-E2. It thus indicates that the current state-of-the-art model still has vulnerability against the NDD attack.

\subsection{Additional Evaluation on More Class Categories}

\begin{figure}[t!]
\centering
\includegraphics[width=\linewidth]{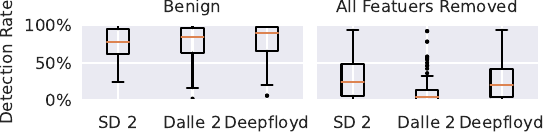}
\vspace{-0.26in}
\caption{Box plots of detection rates for 15 object category images generated by the 3 diffusion models.}
\label{fig:box_other_cat}
\vspace{-0.15in}
\end{figure}

Fig.~\ref{fig:box_other_cat} shows box plots of the detection rates of the 15 object categories (stop sign, car, dog, hot dog, traffic light, zebra, fire hydrant, frog, horse, bird, boat, air plane, bicycle, cat, and carrot). We evaluate 6 object detectors: YOLOv3, YOLOv5, DETR, Faster R-CNN, RTMDet, and YOLOv8. Thus, there are 90 (15 $\times$ 6) data points for each plot.
As shown, there are always some levels of vulnerability against the NDD attack because the median values are above zero. Our paper covered the general classes of the object: a stop sign for an artificial sign, a fire hydrant for an artificial object, and a horse for a natural object. We thus leave a deep analysis of other categories for the following works.

\nsection{Detailed Results of Natural Attack Capability on Image Classification Models}

\textit{Experimental setup.}
We train 4 state-of-the-art image classification models, ResNet50~\cite{he2016deep}, DenseNet121~\cite{huang2017densely}, EfficientNet b0~\cite{tan2019efficientnet}, and ResNeXt~\cite{xie2017aggregated}, on the training datasets derived from the COCO dataset~\cite{lin2014microsoft}. We convert the COCO dataset into a dataset for the multiclass classification task in the same way as we did in RQ4: We crop the images in the COCO dataset with their bounding box annotations and randomly select 500 images for each class.

\textit{Results.} Table~\ref{tbl:classification_stop_signs},~\ref{tbl:classification_fire_hydrants}, and~\ref{tbl:classification_horse} show the classification accuracy for the 3 classes: stop signs, fire hydrants, and horses.
As shown, the large number of the generated images are still classified as the targeted object class even though we remove a robust feature. For stop sign, $\geq$47\% of the generated images are still classified as stop signs even though all 4 robust features are removed. For fire hydrant and horse, $\geq$38\% and $\geq$16\% of the generated images are classified as the original class, respectively. In summary, the NDD attack shows quite high attack effectiveness against not only object detectors but also image classifiers.

\begin{table}[h!]
\footnotesize
\centering
\renewcommand{\arraystretch}{1.3}
\setlength{\tabcolsep}{1pt}
\captionof{table}{Classification accuracy of 4 classifiers on the \textbf{stop sign} images in the NDDA dataset generated by the 3 diffusion models. \textbf{Bold} and \underline{underline} denote highest and lowest scores in each row.
}
\label{tbl:classification_stop_signs}
\vspace{-0.1in}
\begin{tabular}{ccccc|cccc|cc}
\toprule
                                    & \multicolumn{4}{c|}{\begin{tabular}[c]{@{}c@{}}Removed \\ Robust Features\end{tabular}} & \multicolumn{4}{c|}{Image Classifiers}            &       \\ \cline{2-10}
                                    & Shape     & Color & Text  & Pattern    & Resnet50 & Dense121 & Effic.Net & Resnext & Avg.\\ \hline
\multirow{6}{*}{\rotatebox{90}{DALL-E 2}}      &     &    &     &      & \textbf{100\%}  & \textbf{100\%}   & \textbf{100\%} & \textbf{100\%}  & 100\% \\
                                    &   \CheckmarkBold      &   &    &       & \textbf{98\%}   & \textbf{98\%}   & \textbf{98\%}  & \textbf{98\%} & 98\%    \\
                                    &    &   \CheckmarkBold   & &   & \textbf{100\%}   & \textbf{100\%}   & \textbf{100\%}  & \textbf{100\%}  & 100\%     \\
                                    &  &   &  \CheckmarkBold   & & \underline{94\%}  & \underline{94\%}  & \underline{94\%}  & \underline{96\%}  & 95\%   \\
                                    &  &   & &  \CheckmarkBold   & \underline{84\%}   & 88\%  & 86\%  & \textbf{90\%}  & 87\%   \\
                                    &  \CheckmarkBold       &  \CheckmarkBold    &   \CheckmarkBold  & \CheckmarkBold  & 42\%   & \underline{58\%}    & 40\%   & \textbf{64\%}   & 51\%  \\
                                    \hline
\multirow{6}{*}{\rotatebox{90}{\hspace{1em}Stable Diffusion 2}} &       &      &    &  & 72\%   & \textbf{76\%}   & \underline{60\%} & 66\%    & 69\%     \\
                                    &   \CheckmarkBold      &   &    &       & 50\%   & 56\%   & \underline{40\%}  & \textbf{80\%}   & 57\%  \\
                                    &    &   \CheckmarkBold   & &   & \textbf{68\%}   & 58\%   & \underline{54\%}  & \textbf{68\%}    & 62\%  \\
                                    &  &   &  \CheckmarkBold &  & 72\%   & 66\%  & \underline{38\%}  & \textbf{76\%}    & 63\%  \\
                                    &  &   & &  \CheckmarkBold   & \textbf{58\%}   & 56\%  & \underline{44\%}  & \textbf{58\%}  & 54\%    \\
                                    &  \CheckmarkBold       &  \CheckmarkBold    &   \CheckmarkBold & \CheckmarkBold    & \textbf{56\%}   & 44\%    & \underline{34\%}   & 54\%   & 47\%  \\
                                    \hline
\multirow{6}{*}{\rotatebox{90}{Deepfloyd IF}}       &    &       &     &     & \textbf{100\%}  & \underline{96\%}   & \textbf{100\%} & \textbf{100\%}    & 99\%      \\
                                                                       &   \CheckmarkBold      &   &    &       & \textbf{96\%}   & \underline{70\%}   & 84\%  & 92\%   & 86\%  \\
                                    &    &   \CheckmarkBold   & &   & \textbf{100\%}   & \underline{86\%}   & \textbf{100\%}  & \textbf{100\%}   & 97\%   \\
                                    &  &   &  \CheckmarkBold &  & 92\%  & 86\%  & \underline{74\%} & \textbf{96\%}     & 87\% \\
                                    &  &   & &  \CheckmarkBold   & 88\%   & 86\%  & \underline{78\%}  & \textbf{92\%}  & 86\%    \\
                                    &  \CheckmarkBold       &  \CheckmarkBold    &   \CheckmarkBold & \CheckmarkBold & \textbf{90\%}   & 86\%    & \underline{52\%}   & \textbf{90\%}  & 78\%   \\
                                    \hline
\end{tabular}
\end{table}

\begin{table}[h!]
\footnotesize
\centering
\renewcommand{\arraystretch}{1.3}
\setlength{\tabcolsep}{2pt}
\captionof{table}{Classification accuracy of 4 classifiers on the \textbf{fire hydrant} images in the NDDA dataset generated by the 3 diffusion models. \textbf{Bold} and \underline{underline} denote highest and lowest scores in each row.
}
\label{tbl:classification_fire_hydrants}
\vspace{-0.1in}
\begin{tabular}{cccc|cccc|cc}
\toprule
                                    & \multicolumn{3}{c|}{\begin{tabular}[c]{@{}c@{}}Removed \\ Robust Features\end{tabular}} & \multicolumn{4}{c|}{Image Classifiers}            &      \\ \cline{2-10}
                                    & Shape     & Color     & Pattern    & Resnet50 & Dense121 & Effic.Net & Resnext & Avg. \\ \hline
\multirow{6}{*}{\rotatebox{90}{DALL-E 2}}           &    &     &      & 96\%  & \textbf{98\%}   & \underline{84\%} & 96\% & 94\%  \\
                                    &   \CheckmarkBold      &       &       & \textbf{94\%}   & 86\%   & 78\%  & \underline{76\%}  & 84\%  \\
                                    &    &   \CheckmarkBold   &    & \textbf{94\%}   & \underline{88\%}   & 90\%  & 90\%   & 91\%  \\
                                    &  &   &  \CheckmarkBold   & \textbf{76\%}   & 62\%  & \underline{36\%}  & 52\%   & 57\%  \\
                                    &  \CheckmarkBold       &  \CheckmarkBold    &   \CheckmarkBold    & \textbf{66\%}   & 62\%    & 64\%   & \underline{46\%} & 60\%    \\
                                    \hline
\multirow{6}{*}{\rotatebox{90}{\hspace{1em}Stable Diffusion 2}} &       &      &      & \textbf{90\%}   & \textbf{90\%}   & \underline{78\%} & 84\%    & 86\%     \\
                                    &   \CheckmarkBold      &      &       & 50\%   & 46\%   & \underline{40\%}  & \textbf{52\%}    & 47\%      \\
                                    &    &   \CheckmarkBold    &     & \textbf{94\%}   & \underline{86\%}   & 88\%  & \underline{86\%}      & 89\%      \\
                                    & &   &  \CheckmarkBold   & 68\%   & 64\%   & \underline{62\%}  & \textbf{72\%}   & 67\%      \\
                                    &  \CheckmarkBold    &   \CheckmarkBold   &   \CheckmarkBold    & \textbf{46\%}   & \underline{26\%}    & 34\%   & \textbf{46\%}    & 38\%       \\
                                    \hline
\multirow{6}{*}{\rotatebox{90}{Deepfloyd IF}}       &    &       &          & \underline{98\%}  & \underline{98\%}   & \textbf{100\%} & \textbf{100\%}   & 99\%       \\
                                    &   \CheckmarkBold      &       &       & \textbf{94\%}   & \underline{90\%}   & \textbf{94\%}  & \textbf{94\%} & 93\%   \\
                                    &    &   \CheckmarkBold    &     & 94\%   & \underline{92\%}   & 96\%  & \textbf{100\%} & 96\%  \\
                                    & &    &  \CheckmarkBold   & \textbf{96\%}   & 88\%   & \underline{82\%}  & 90\% & 89\%        \\
                                    &  \CheckmarkBold       &  \CheckmarkBold   &   \CheckmarkBold    & 86\%   & 82\%    & \underline{74\%}   & \textbf{98\%}   & 85\%   \\
                                    \hline
\end{tabular}
\end{table}

\begin{table}[h!]
\footnotesize
\centering
\renewcommand{\arraystretch}{1.3}
\setlength{\tabcolsep}{2pt}
\captionof{table}{Classification accuracy of 4 classifiers on the \textbf{horse} images in the NDDA dataset generated by the 3 diffusion models. \textbf{Bold} and \underline{underline} denote highest and lowest scores in each row.
}
\label{tbl:classification_horse}
\vspace{-0.1in}
\begin{tabular}{cccc|cccc|cc}
\toprule
                                    & \multicolumn{3}{c|}{\begin{tabular}[c]{@{}c@{}}Removed \\ Robust Features\end{tabular}} & \multicolumn{4}{c|}{Image Classifiers}            &      \\ \cline{2-10}
                                    & Shape     & Color     & Pattern    & Resnet50 & Dense121 & Effic.Net & Resnext & Avg. \\ \hline
\multirow{6}{*}{\rotatebox{90}{DALL-E 2}}           &    &     &      & 64\%  & 60\%   & \underline{48\%} & \textbf{66\%}  & 60\%  \\
                                    &   \CheckmarkBold      &       &       & 80\%   & \textbf{84\%}   & \underline{74\%}  & 82\% & 80\%    \\
                                    &    &   \CheckmarkBold   &    & 34\%   & 32\%   & \underline{18\%}  & \textbf{38\%}  & 31\%    \\
                                    &  &   &  \CheckmarkBold   & \textbf{18\%}   & 6\%  & \underline{0\%}  & 6\%  & 8\%    \\
                                    &  \CheckmarkBold       &  \CheckmarkBold    &   \CheckmarkBold    & \textbf{26\%}   & 14\%    & \underline{10\%}   & 12\% & 16\%    \\
                                    \hline
\multirow{6}{*}{\rotatebox{90}{\hspace{1em}Stable Diffusion 2}} &       &      &      & \textbf{90\%}   & 78\%   & \underline{74\%} & 82\%   & 81\%      \\
                                    &   \CheckmarkBold      &      &       & 92\%   & \textbf{98\%}   & \underline{86\%}  & 94\%    & 93\%      \\
                                    &    &   \CheckmarkBold    &     & \textbf{80\%}   & 54\%   & \underline{40\%}  & 54\%      & 57\%      \\
                                    & &   &  \CheckmarkBold   & \textbf{32\%}   & 24\%   & \underline{14\%}  & 28\%   & 25\%      \\
                                    &  \CheckmarkBold    &   \CheckmarkBold   &   \CheckmarkBold    & \underline{46\%}   & \textbf{62\%}    & 48\%   & 54\%    & 53\%       \\
                                    \hline
\multirow{6}{*}{\rotatebox{90}{Deepfloyd IF}}       &    &       &          & \textbf{94\%}  & 92\%   & \underline{82\%} & \textbf{94\%}    & 61\%      \\
                                    &   \CheckmarkBold      &       &       & 90\%   & \textbf{92\%}   & 86\%  & \underline{82\%} & 88\%   \\
                                    &    &   \CheckmarkBold    &     & \textbf{94\%}   & 90\%   & \underline{80\%}  & 86\% & 88\%  \\
                                    & &    &  \CheckmarkBold   & 60\%   & \textbf{64\%}   & \underline{44\%}  & 50\%  &   55\%    \\
                                    &  \CheckmarkBold       &  \CheckmarkBold   &   \CheckmarkBold    & 52\%   & \textbf{62\%}    & \underline{34\%}   & \underline{34\%}  & 46\%    \\
                                    \hline
\end{tabular}

\end{table}

\nsection{Detailed Results of GAN-based Attacks (RQ1)}

\begin{figure}[h!]
\centering
\includegraphics[width=\linewidth]{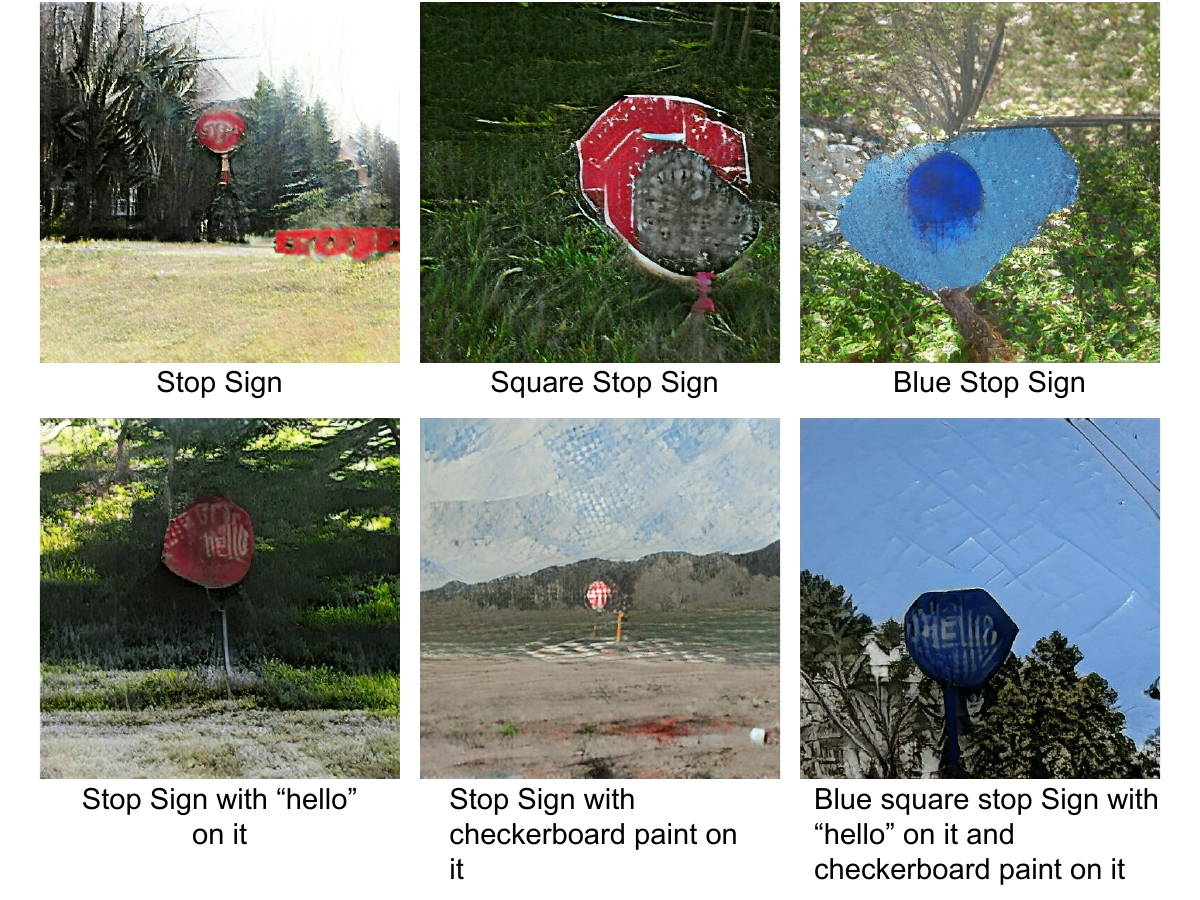}
\vspace{-0.2in}
\caption{Examples of stop sign images generated by BigSleep that successfully trick at least one object detector, i.e., a stop sign is detected with a confidence score $\geq$ 0.5. The user survey includes these images and more.}
\label{fig:biggan_stop_signs}
\vspace{-0.2in}
\end{figure}

Fig.~\ref{fig:biggan_stop_signs} shows the attacks generated by BigSleep that successfully trick object detectors with a confidence score $\geq$ 0.5. As shown, GAN-based NDD attacks can still generate several successful attacks which have high stealthiness as confirmed by the user study. While the diffusion models have much higher attack effectiveness as discussed in RQ1, this natural attack capability in the GAN model can be also a serious attack threat.
Considering such a high attack capability has not been reported even though many GAN-based adversarial attacks~\cite{bai2021ai} are proposed, we think this is mainly due to the high-quality text guide by OpenAI CLIP~\cite{radford2021learning} in BigSleep~\cite{bigsleep} rather than due to a unique characteristic of the GAN model. However, it is not trivial to investigate the relationship between the impact of non-robust features and the quality of the text guide. We hope that a future study can handle this. So far, the DDA should be more preferable to the attackers because BigSleep's image generation takes much longer ($\geq$20 minutes) than the diffusion model (a few seconds), even though BigSleep needs to generate more images due to its low attack success rate.

\nsection{Detailed Setup of the User Study (RQ2)}

To conduct an ethical user study, we have gone through the IRB process in our institution. The entire questionnaire form of this user study is attached as a part of the supplementary materials.
We recruited 82 human subjects on Prolific~\cite{prolific}, a crowdsourcing platform specialized for research purposes.
All human subjects are English speakers in the United States to follow the IRB instruction. We did not collect any other demographic or privacy-sensitive information from the participants.

We provide 3 types of images: benign, diffusion-generated, and GAN-generated. 3 benign images are provided first in order to have a set of baseline results. Then, we show diffusion-generated images to the users; for every text prompt used, 3 images are generated. A text prompt may either produce a benign image, an image with 1 robust feature removed, or an image with all robust features removed. This step is also repeated using the BigSleep~\cite{bigsleep} model.
Fig.~\ref{fig:biggan_stop_signs} shows examples of stop sign images generated by BigSleep that successfully trick at least one object detector, i.e., a stop sign is detected with a confidence score $\geq$0.5. The user survey includes these images and more.

\nsection{Additional Results of the Experiment on Non-Robust Features (RQ4)}

Table~\ref{tbl:firehydrant_robust_classifier} and~\ref{tbl:horse_robust_classifier} show the accuracy of the robust and normal classifiers on
the fire hydrant and horse object classes, respectively. For the fire hydrant images, the robustified classifier drops the accuracy when the robust features are removed, as observed in RQ4. For the horse images, this analysis does not show meaningful results because the training of robustified classifier fails as the accuracy is $<$40\% even for the benign images. We manually check the robustified horse images and find that the robustified images do not look like legitimate horses. Since this methodology~\cite{ilyas2019adversarial} is originally evaluated on the CIFAR-10~\cite{cifar10}, it may not be fully compatible with the dataset derived from the COCO dataset.
Nevertheless, our finding is generally observed in other cases when the robustified classifiers can achieve similar accuracy to the normal classifier in the benign images.

\begin{table}[h!]
\centering
\footnotesize
\setlength{\tabcolsep}{3.1pt}
\setlength{\aboverulesep}{0pt}
\setlength{\belowrulesep}{0pt}
\caption{Accuracy of the robust and normal classifiers on the \textbf{fire hydrant} images in the NDDA dataset. BThe benign means the images generated with the benign prompts; The NDD means the NDD attack that removes all robust features.}
\vspace{-0.1in}
\label{tbl:firehydrant_robust_classifier}
\begin{tabular}{cccc||ccc}
\bottomrule
     & \multicolumn{3}{c||}{Robustified classifier} & \multicolumn{3}{c}{Normal classifier}      \\ \cline{2-7}
     & Benign  & \multicolumn{1}{c|}{NDD}  & Diff. & Benign & \multicolumn{1}{c|}{NDD}  & Diff. \\ \hline
DALL-E 2           & 0.66 & \multicolumn{1}{c|}{0.1} & 0.56 & 0.94 & \multicolumn{1}{c|}{0.36} & 0.58 \\
Stable Diffusion 2 & 0.96 & \multicolumn{1}{c|}{0.42} & 0.54 & 0.96 & \multicolumn{1}{c|}{0.38} & 0.58 \\
DeepFloyd IF       & 0.82 & \multicolumn{1}{c|}{0.2} & 0.62 & 1.00 & \multicolumn{1}{c|}{0.9} & 0.1  \\ \hline
Avg. & 0.81    & \multicolumn{1}{c|}{0.24} & \textbf{0.57}  & 0.97 & \multicolumn{1}{c|}{0.55} & \textbf{0.39}  \\ \bottomrule
\end{tabular}
    \vspace{-0.2in}
 \end{table}

\begin{table}[h!]
\centering
\footnotesize
\setlength{\tabcolsep}{3.1pt}
\setlength{\aboverulesep}{0pt}
\setlength{\belowrulesep}{0pt}
\caption{Accuracy of the robust and normal classifiers on the \textbf{horse} images in the NDDA dataset. BThe benign means the images generated with the benign prompts; The NDD means the attack that removes all robust features. Robustified classifier fails to train on the robustified data as its detection rate is $\leq$40\% on the benign images}
\vspace{-0.1in}
\label{tbl:horse_robust_classifier}
\begin{tabular}{cccc||ccc}
\bottomrule
     & \multicolumn{3}{c||}{Robustified classifier} & \multicolumn{3}{c}{Normal classifier}      \\ \cline{2-7}
     & Benign  & \multicolumn{1}{c|}{NDD}  & Diff. & Benign & \multicolumn{1}{c|}{NDD}  & Diff. \\ \hline
DALL-E 2           & 0.10 & \multicolumn{1}{c|}{0.06} & 0.04 & 0.54 & \multicolumn{1}{c|}{0.04} & 0.50 \\
Stable Diffusion 2 & 0.28 & \multicolumn{1}{c|}{0.28} & 0.00 & 0.64 & \multicolumn{1}{c|}{0.24} & 0.40 \\
DeepFloyd IF       & 0.40 & \multicolumn{1}{c|}{0.28} & 0.12 & 0.9 & \multicolumn{1}{c|}{0.28} & 0.62 \\ \hline
Avg. & 0.26    & \multicolumn{1}{c|}{0.21} & \textbf{0.05}  & 0.69   & \multicolumn{1}{c|}{0.19} & \textbf{0.51}  \\ \bottomrule
\end{tabular}
    \vspace{-0.2in}
\end{table}

\nsection{Additional Results of Tesla Experiments (RQ6)}

\begin{figure}[h!]
\vspace{-0.1in}
    \centering
    \includegraphics[width=\linewidth]{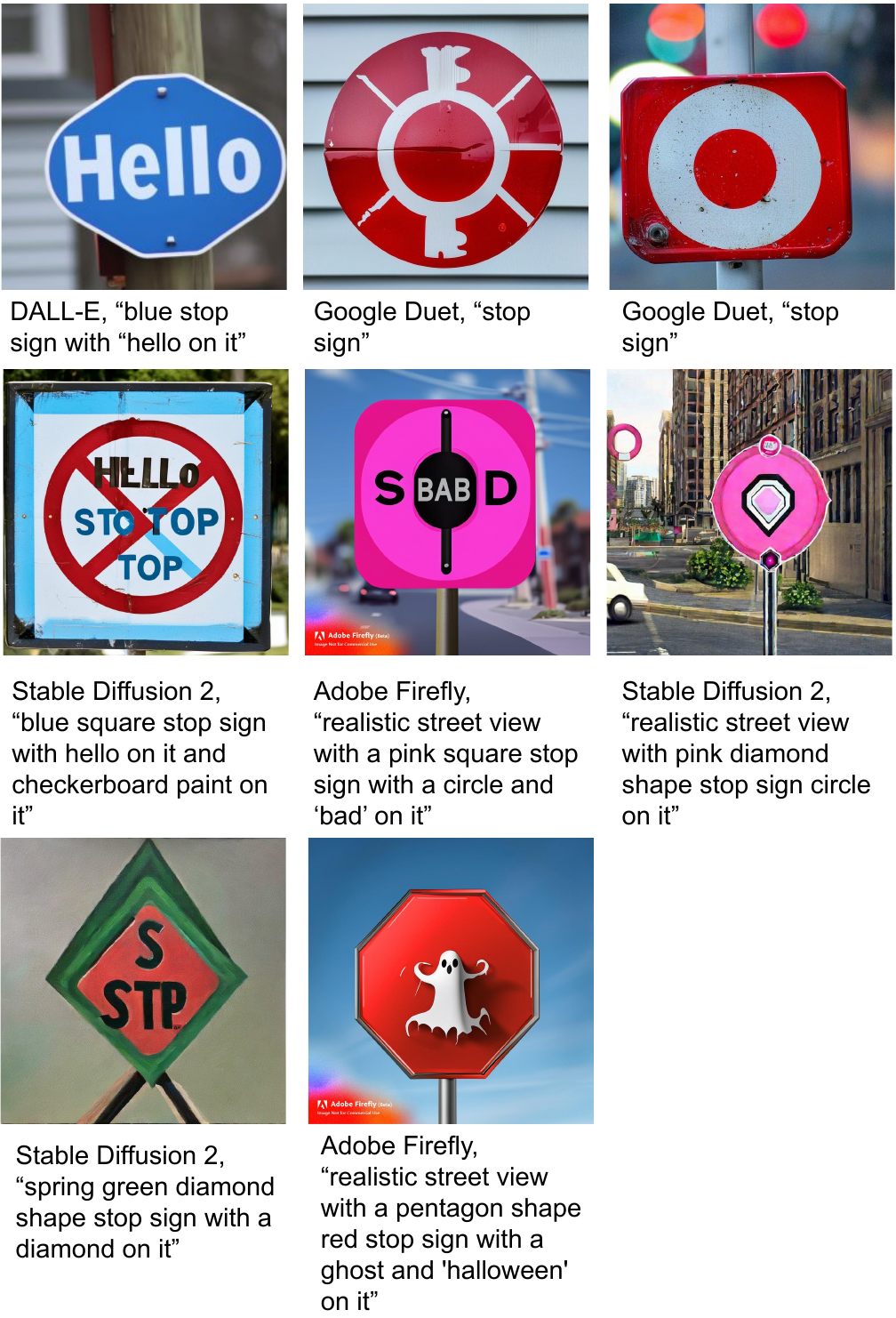}
    \vspace{-0.25in}
    \caption{Successful NDD Attacks on Tesla Model 3. The caption of each image is the used diffusion model and the text prompt.}
    \label{fig:successful_attacks}
    \vspace{-0.3in}
\end{figure}

\begin{figure}[h!]
    \centering
    \includegraphics[width=\linewidth]{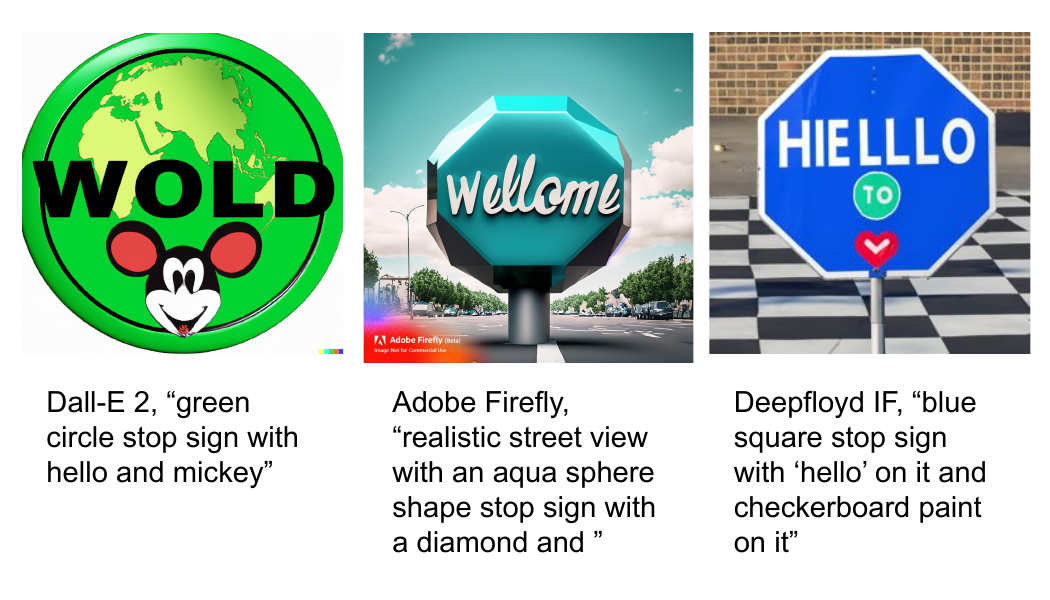}
    \vspace{-0.35in}
    \caption{Unsuccessful NDD Attacks on Tesla Model 3.The caption of each image is the used diffusion model and the text prompt.}
    \vspace{-0.2in}
    \label{fig:unsuccessful_attacks}
\end{figure}

Fig.~\ref{fig:successful_attacks} and~\ref{fig:unsuccessful_attacks} show the stop signs that successfully and unsuccessfully deceived Tesla's vision system respectively. We not only demonstrate that 73\% of these generated images are successful but also show 3/4 of the diffusion models exhibit the natural attack capability in the real world. Out of the models that were successful, most of them required a carefully designed text prompt in order to achieve a successful attack, as illustrated by the last 5 images of Fig.~\ref{fig:successful_attacks} for Adobe Firefly and Stable Diffusion 2. Other diffusion models, such as Google Duet, require less effort; a simple text prompt, ``stop sign'', is enough to generate images that do not look like stop signs but have enough non-robust features to fool object detectors. We plan to inform this vulnerability to Tesla if this paper is accepted. We hope that our study can facilitate further research to train more robust DNN models.

\end{document}